\documentclass[runningheads]{llncs}
\usepackage{graphicx}
\usepackage{amsfonts}       
\usepackage{amsmath}
\usepackage{comment}
\usepackage{subfigure}
\usepackage{booktabs}  
\usepackage{hyperref}
\usepackage{color}
\usepackage{algpseudocode}
\usepackage[vlined,ruled,linesnumbered]{algorithm2e}
\usepackage[misc]{ifsym}
\usepackage{bm}
\usepackage{bbding}
\usepackage{amsmath}
\usepackage[marginal]{footmisc}

\usepackage[marginal]{footmisc}
\usepackage{xspace}
\usepackage{tikz}

\begin{document}
\title{Uncertainty-aware Low-Rank $Q$-Matrix Estimation for Deep Reinforcement Learning }
%
%
\author{Tong Sang \and
Hongyao Tang \and 
Jianye Hao\thanks{Jianye Hao and Yan Zheng are the corresponding authors.} \and
Yan Zheng\textsuperscript{$\small{*}$} \and
Zhaopeng Meng}

\institute{College of Intelligence and Computing, Tianjin University, China
\email{\{2019218044,bluecontra,jianye.hao,yanzheng,mengzp\}@tju.edu.cn}
}

%
%
\maketitle              
\begin{abstract}
\footnote{This paper is accepted by The 3rd International Conference on Distributed Artificial Intelligence (DAI 2021, Shanghai, China).}
Value estimation is one key problem  in Reinforcement Learning.
Albeit many successes have been achieved by Deep Reinforcement Learning (DRL) in different fields, the underlying structure and learning dynamics of value function, especially with complex function approximation, are not fully understood.
In this paper, we report that decreasing rank of $Q$-matrix widely exists during learning process across a series of continuous control tasks for different popular algorithms.
We hypothesize that the low-rank phenomenon indicates the common learning dynamics of $Q$-matrix from stochastic high dimensional space to smooth low dimensional space.
Moreover, we reveal a positive correlation between value matrix rank and value estimation uncertainty.
Inspired by above evidence, we propose a novel \textbf{U}ncertainty-\textbf{A}ware \textbf{L}ow-rank \textbf{$Q$}-matrix \textbf{E}stimation (\textbf{UA-LQE}) algorithm
as a general framework to facilitate the learning of value function.
Through quantifying the uncertainty of state-action value estimation, we selectively erase the entries of highly uncertain values in state-action value matrix 
and conduct low-rank matrix reconstruction for them to recover their values.
Such a reconstruction exploits the underlying structure of value matrix to improve the value approximation, thus leading to a more efficient learning process of value function.
In the experiments, we evaluate the efficacy of UA-LQE in several representative OpenAI MuJoCo continuous control tasks.
\keywords{Reinforcement Learning  \and Value Estimation \and Uncertainty \and Low Rank.}
\end{abstract}

\section{Introduction}
\label{section:introduction}
In recent years, Deep Reinforcement Learning (DRL) has been demonstrated to be a promising approach to solve complex sequential decision-making problems
in different domains \cite{Lillicrap2015DDPG,Mnih2015DQN,Silver2016Go,vinyals2019grandmaster,HafnerLB020Dream,schreck2019retrosyn}.
Albeit the progress made in RL, the intriguing learning dynamics is still not well known, especially when with complex function approximation, e.g., deep neural network (DNN).
The little understanding of how RL performs and proceeds prevents the deep improvement and utilization of RL algorithms.

To understand the learning dynamics of RL,
value function is one of the most important functions to consider.
Value function defines the expected cumulative rewards (i.e., returns) of a policy,
indicating how a state or taking an action under a state could be beneficial when performing the policy.
It plays a vital role in RL 
for both value-based methods (e.g., DQN \cite{Mnih2015DQN}) 
and policy-based methods (e.g., DDPG \cite{Lillicrap2015DDPG}).
The learning dynamics and underlying structure of value function can be easily investigated in tabular RL by classical RL algorithms \cite{Sutton1988ReinforcementLA}, where typically the action-value function $Q$ of finite states and actions is maintained as a table.
Towards larger state-action space,
linear function approximation is widely as a representative option used for analysis \cite{ScherrerGGLG15AMDP,LyleROD21OntheE,He21UniformPAC}.
Nevertheless, with complex non-linear function approximation as DRL, the learning dynamics and underlying structure are not well understood.
A few works provide some insights on this with different empirical discoveries \cite{Hasselt18DRLDT,YangZXK20SVRL,LuoMHCW20I4R,KumarAGL21ImplicitUnder}.

One notable work is Structured Value-based RL (SVRL) \cite{YangZXK20SVRL}.
Yang et al. \cite{YangZXK20SVRL} identifies the existence of low-rank $Q$-functions 
in both classical control with Dynamic Programming \cite{Ong15lowrank} and Atari games with DQN \cite{Mnih2015DQN}.
In specific, 
the approximate rank of $Q$-matrix 
spanned by (sampled) states and actions
decreases during the learning process.
This indicates an underlying low-rank structure of the optimal value function and the learning dynamics of rank decrease.
By leveraging this low-rank structure of $Q$-function,
SVRL performs random drop of $Q$-matrix entries and then matrix estimation (or matrix completion).
The intuition behind SVRL is the utilization of global structure of $Q$-function to reduce the value estimation errors thus facilitates the value function learning.
Despite of the promising results achieved by SVRL,
the low-rank structure of value function has not been investigated in continuous control.
In addition, the selection of $Q$-matrix entries is aimless,
in other words, it does not take into account the quality of estimate values,
thus the underlying information from the low-rank structure is not fully and accurately utilized.

\begin{figure}[t]
\centering
\hspace{-0.55cm}
\subfigure[Hopper]{
\includegraphics[width=0.34\textwidth]{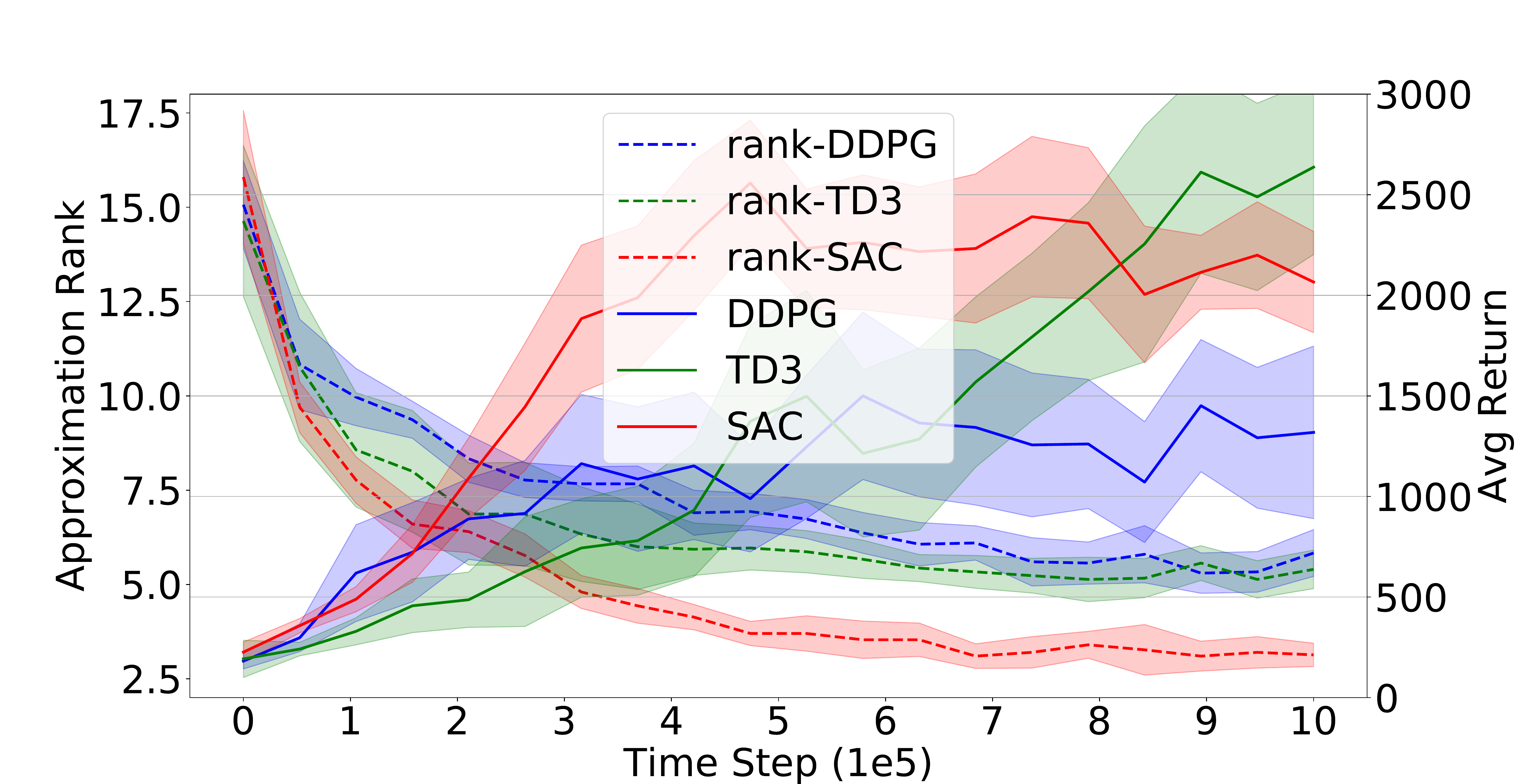}
}
\hspace{-0.45cm}
\subfigure[Walker2d]{
\includegraphics[width=0.34\textwidth]{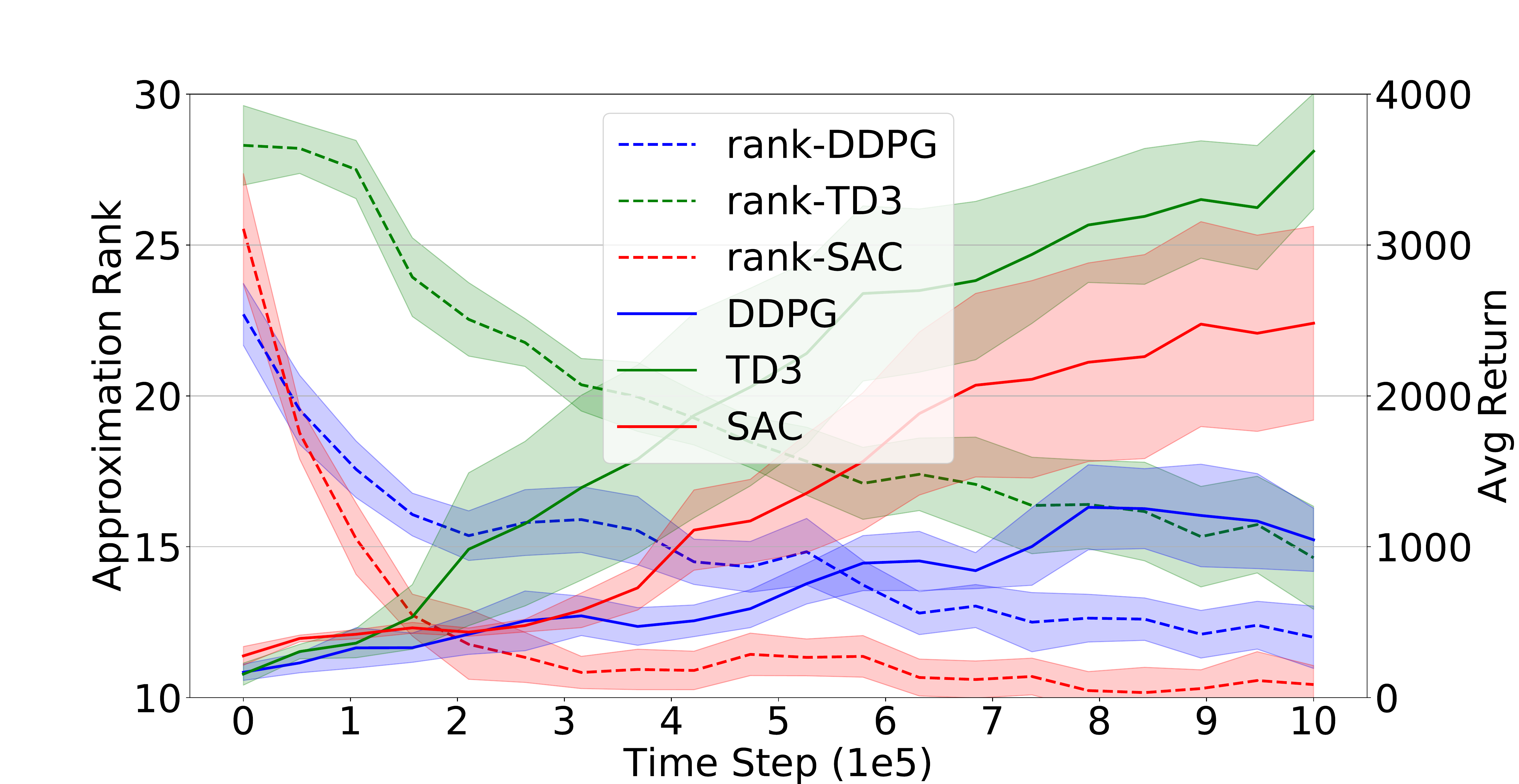}
}
\hspace{-0.45cm}
\subfigure[Ant]{
\includegraphics[width=0.34\textwidth]{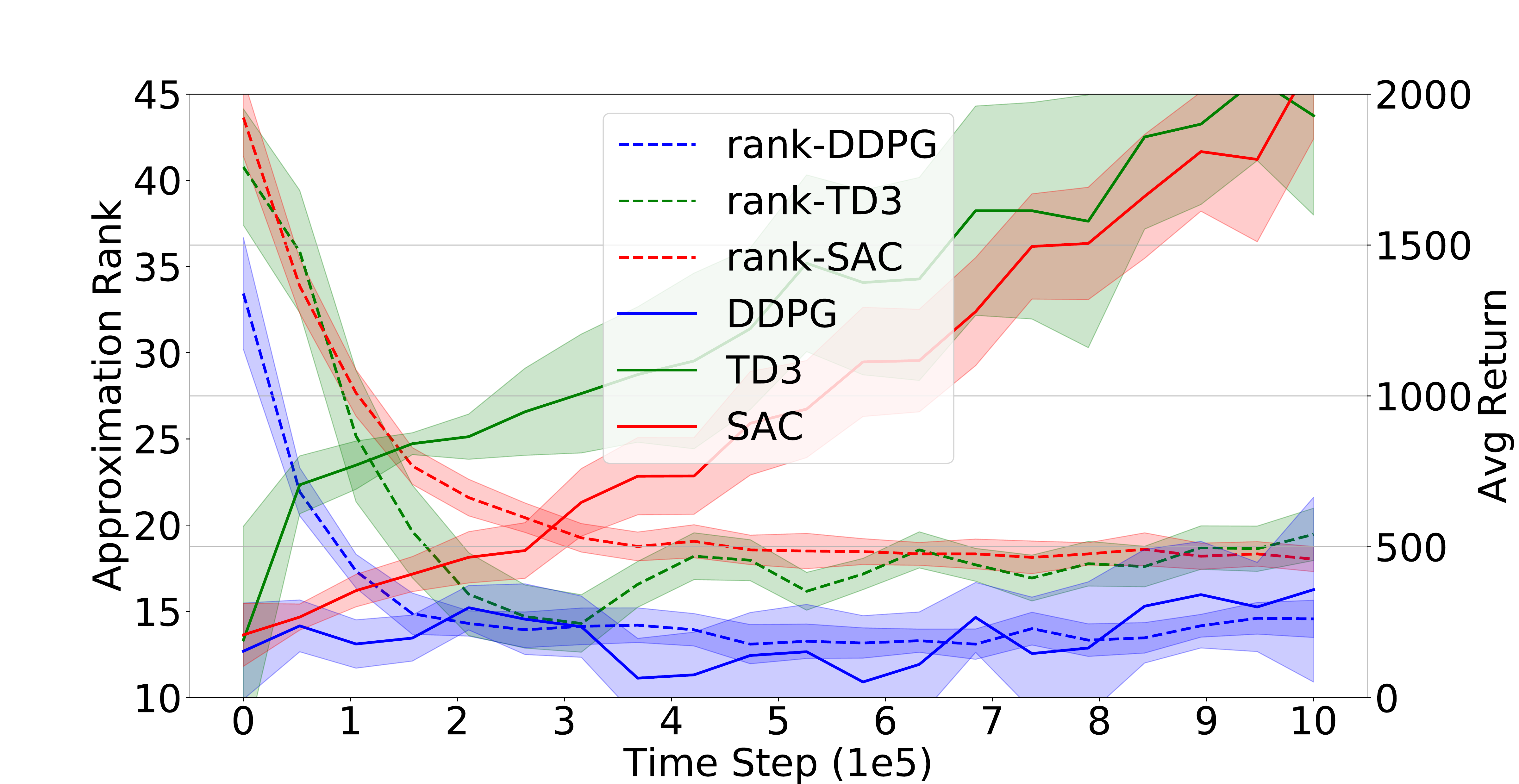}
}
\vspace{-0.4cm}
\caption{
\textbf{Low-rank structure of $Q$-matrix} in MuJoCo continuous control tasks.
The left and right vertical axes denote the approximate rank of $Q$-matrix and average return.
For all tasks and all algorithms,
the rank decreases along the learning process.
The results are averaged over 3 runs and the shaded region denotes a standard deviation.
}
\label{figure:low_rank_ph}
\end{figure}

In this paper,
we aim at addressing the above two problems and propose a novel algorithm called \textbf{U}ncertainty-\textbf{A}ware \textbf{L}ow-rank \textbf{$Q$}-matrix \textbf{E}stimation (\textbf{UA-LQE}),
as a general framework to facilitate the learning of value function of DRL agents.
We first examine the low-rank structure and learning dynamics of $Q$-functions in continuous control task.
A quick view of the conclusion is shown in Fig. \ref{figure:low_rank_ph},
the rank of sampled $Q$-matrix of representative continuous control DRL algorithms (DDPG \cite{Lillicrap2015DDPG}, TD3 \cite{Fujimoto2018TD3} and SAC \cite{HaarnojaZAL18SAC}) decreases during the learning process.
Our empirical evidence completes the SVRL's discovery in DRL,
indicating the generality of low-rank structure of $Q$-function across different algorithms and environments.
Thus, we can expect that an effective method that makes use of the low-rank structure to improve RL will be beneficial in RL community broadly.
Moreover,
we take a step further and empirically reveal a positive correlation between value matrix rank and value estimation uncertainty,
i.e., a value matrix of high rank often has an high overall value estimation uncertainty.
Thus, we hypothesize that value estimation uncertainty can be used as the indicator of the target entries in $Q$-value matrix.
This is because that 
the value estimate with high uncertainty often has a large estimation error 
that may severely violate the underlying structure of $Q$-value matrix,
making themselves be the major `sinners' to the induced high rank of value matrix.
To this end,
rather than erasing the entries and performing matrix estimation to reconstruct the value aimlessly as in SVRL,
we quantify the uncertainty of state-action value estimates,
then reconstruct the entries of high uncertainty in $Q$-value matrix.
We expect the uncertainty-aware reconstruction to better reduce the value estimation errors and facilitate the emergence of the low-rank structured value function.
In specific, we adopt deep ensemble-based and count-based uncertainty quantification
and conduct value matrix reconstruction on both online value estimates (from evaluate network) and target value estimates (from target network).
In the experiments, we evaluate the effectiveness of UA-LQE with DDPG \cite{Lillicrap2015DDPG} as the base algroithm in OpenAI MuJoCo continuous control tasks.

We summarize our contributions as follows:
1) We empirically demonstrate the existence of low-rank structure of $Q$-function for representative DRL algorithms in widely-adopted continuous control tasks.
2) We empirically reveals a positive correlation between value matrix rank and value estimation uncertainty.
3) We propose a novel algorithm, i.e., Uncertainty-Aware Low-rank $Q$-matrix Estimation (UA-LQE).
We demonstrate the effectiveness of UA-LQE in improving learning performance in our experiments. 

\section{Background}
\label{section:background}

\subsection{Reinforcement Learning}
\label{section:bg_rl}
Consider a Markov Decision Process (MDP) defined by the tuple $\langle S, A, P, R, \gamma, T \rangle$ with state space $S \in \mathbb{R}^{d_s}$, action space $A \in \mathbb{R}^{d_a}$,
transition function $P: S \times A \times S \to [0,1]$, reward function $R: S \times A \to \mathbb{R}$, discount factor $\gamma \in [0,1)$
and horizon $T$.
The agent interacts with the MDP by performing its policy $\pi: S \rightarrow A$.
An RL agent aims at optimizing its policy to maximize the expected discounted cumulative reward
$J(\pi) = \mathbb{E}_{\pi} [\sum_{t=0}^{T}\gamma^{t} r_t ]$,
where $s_{0} \sim \rho_{0}\left(s_{0}\right)$ the initial state distribution, $a_{t} \sim \pi\left(s_{t}\right)$, $s_{t+1} \sim \mathcal{P}\left(s_{t+1} \mid s_{t}, a_{t}\right)$ and $r_t = R\left(s_{t},a_{t}\right)$.
The state-action value function $Q^{\pi}$ is defined as 
the expected cumulative discounted reward:
$Q^{\pi}(s, a)=\mathbb{E}_{\pi} \left[\sum_{t=0}^{T} \gamma^{l} r_{t} \mid s_{0}=s, a_{0}=a \right]$
for all $s,a \in S \times A$.


With function approximation, the agent maintains a parameterized policy $\pi_{\phi}$,
while $Q^{\pi_{\phi}}$ can be approximated by $Q_{\theta}$ with parameter $\theta$
typically through minimizing Temporal Difference loss \cite{Sutton1988ReinforcementLA}:
\vspace{-0.1cm}
\begin{equation}
    L(\theta)=\mathbb{E}_{s,a,r,s^{\prime} \sim D} \left[Q_{\theta}(s, a) - \mathbb{E}_{a^{\prime} \sim \pi_{\bar{\phi}}(s^{\prime})} \left(r + \gamma Q_{\bar{\theta}}(s^{\prime},a^{\prime})\right) \right]^{2}.
\label{eq:q_td}
\vspace{-0.1cm}
\end{equation}
The policy $\pi_{\phi}$ can be updated by taking the gradient of the objective $\nabla_{\phi} J(\pi_{\phi})$,
e.g., 
with the deterministic policy gradient (DPG) \cite{Silver2014DPG}:
\vspace{-0.1cm}
\begin{equation}
    \nabla_{\phi} J(\pi_{\phi}) = \mathbb{E}_{s \sim \rho^{\pi_{\phi}}} \left[ \nabla_{\phi} \pi_{\phi}(s) \nabla_{a} Q^{\pi}(s,a)|_{a=\pi_{\phi}(s)}\right],
\label{eq:DPG}
\vspace{-0.1cm}
\end{equation}
where $\rho^{\pi_{\phi}}$ is the discounted state distribution under policy $\pi_{\phi}$.
$\bar{\phi},\bar{\theta}$ are the parameters of \textit{target networks}, which are periodically or smoothly updated from the parameters (i.e., $\phi,\theta$) of \textit{evaluation networks}.

\vspace{-0.15cm}

\subsection{Approximate Rank and Matrix Reconstruction}
\label{section:bg_lr}

For a finite state space and a finite action space,
we can view the corresponding $Q$-function as a $Q$-value matrix (shortly $Q$-matrix) of shape $|S|\times|A|$.
The approximate rank \cite{YangZXK20SVRL,KumarAGL21ImplicitUnder} for a threshold $\delta$ is defined as arank$_{\delta}(Q) = \min \{k:\sum_{i=1}^{k}\sigma_{i}(Q)\ge (1-\delta)\sum_{i=1}^{d}\sigma_{i}(Q)\}$,
where $\{\sigma_{i}(Q)\}$ are the singular values of $Q$-value matrix in decreasing order, i.e., $\sigma_1 \ge \dots \ge \sigma_d \ge 0$.
In this paper, we use $\delta=0.01$ and omit the subscript for clarity.
In other words, approximate rank is the first $k$
singular values that capture more than 99\%  variance of all singular values.
Further, for continuous (infinite) state and action space,
we can define that arank$(Q) = \frac{1}{N}\sum_{j=1}^{N}$arank$(\tilde{Q}_j)$, i.e.,
the empirical average approximate rank of sampled $Q$-value submatrix $\{\tilde{Q}_i\}$, each of which corresponds to sampled state and action subspaces $S_i,A_i$.

If arank$(Q) \ll \min \{|S|,|A|\}$,
we say a $Q$-value matrix (or $Q$-function) is \textit{low-rank}.
This means that the matrix can be \textit{redundant} in the sense that the whole matrix can be reconstructed or completed with only some entries are known.
In this paper, we consider Soft-Impute algorithm \cite{MazumderHT10SoftImpute} for matrix reconstruction and more background about matrix estimation is provided in Appendix \ref{appendix:bg_matrix_estimation}.
In the context of RL, we also view this as the underlying structure of $Q$-value matrix,
of which we make use to improve value function learning in RL.

\section{Low-Rank $Q$-Matrix in DRL}
\label{section:empirical_evidence}

In this section, we first empirically investigate the rank of $Q$-matrix for representative DRL agents in continuous control tasks (Sec.~\ref{subsection:emp_low_rank}),
which complements prior studies in discrete-action control tasks.
Then, 
we introduce two ways to leverage the low-rank structure for DRL continuous control (Sec.~\ref{subsection:matrix_rec}).

\vspace{-0.15cm}

\subsection{Empirical Study of Low-Rank $Q$-Matrix in MuJoCo}
\label{subsection:emp_low_rank}

\begin{figure}[t]
\centering
\hspace{-0.45cm}
\subfigure[Hopper]{
\includegraphics[width=0.345\textwidth]{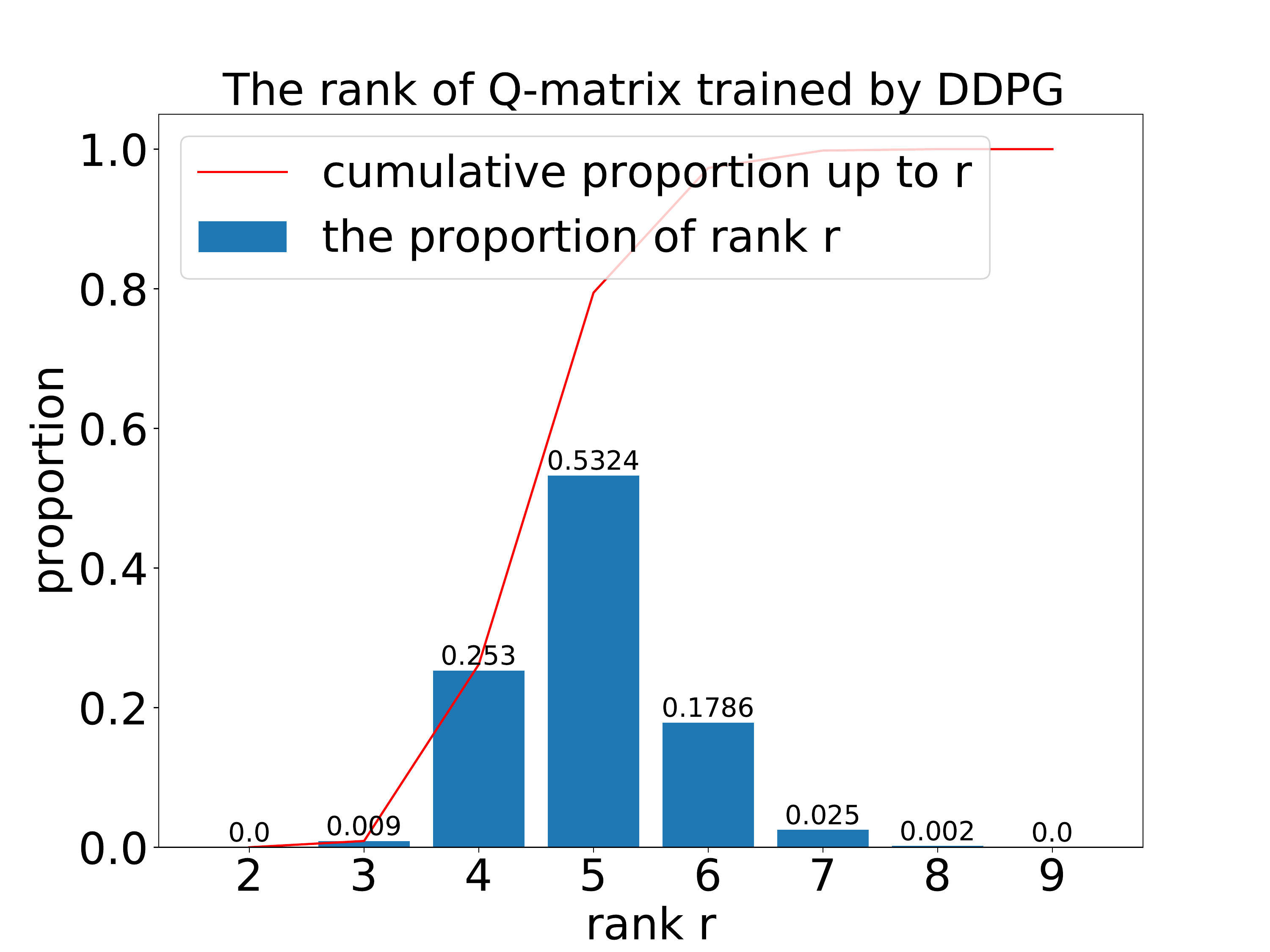}
}
\hspace{-0.68cm}
\subfigure[Walker2d]{
\includegraphics[width=0.345\textwidth]{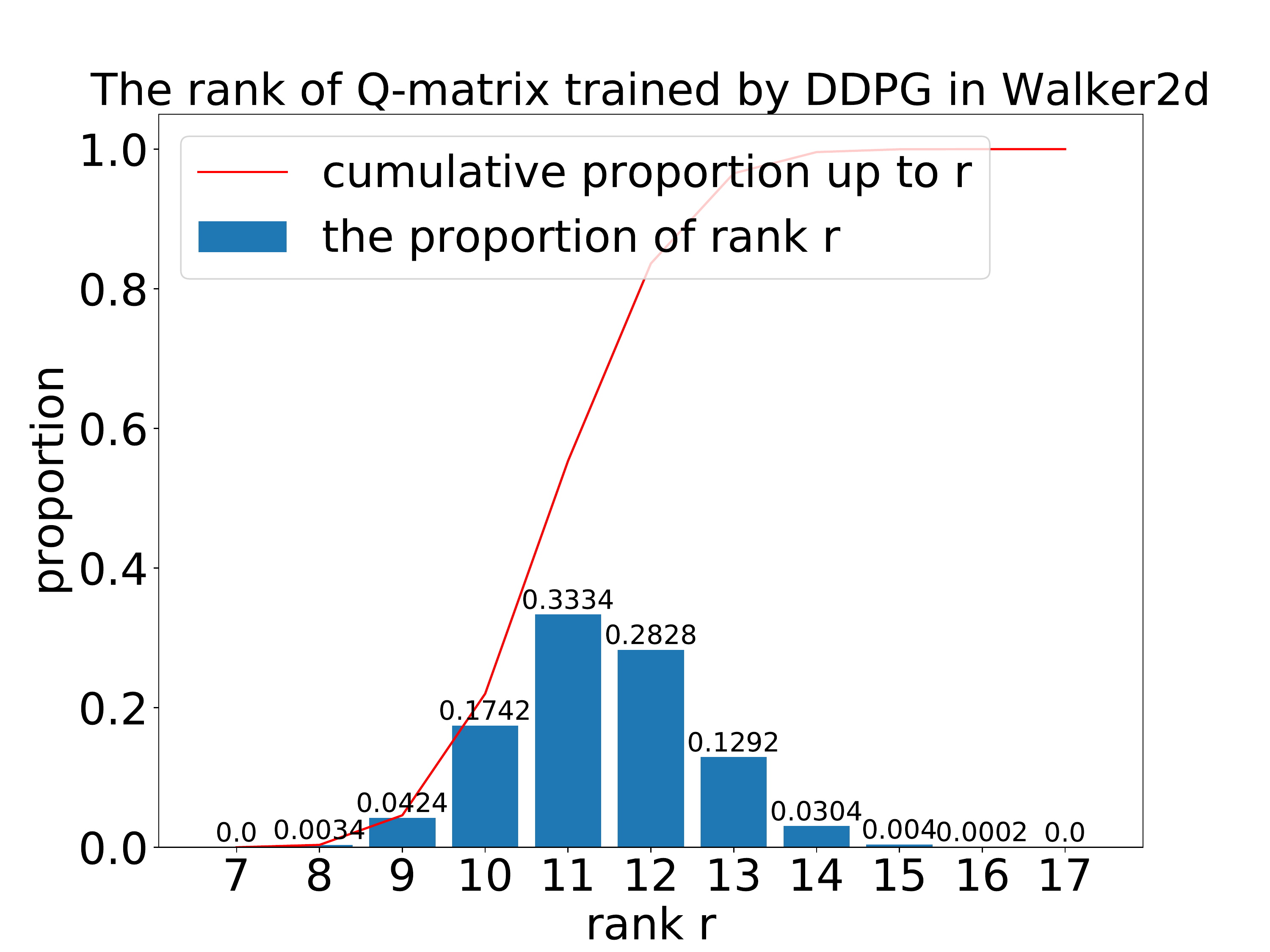}
}
\hspace{-0.68cm}
\subfigure[Ant]{
\includegraphics[width=0.345\textwidth]{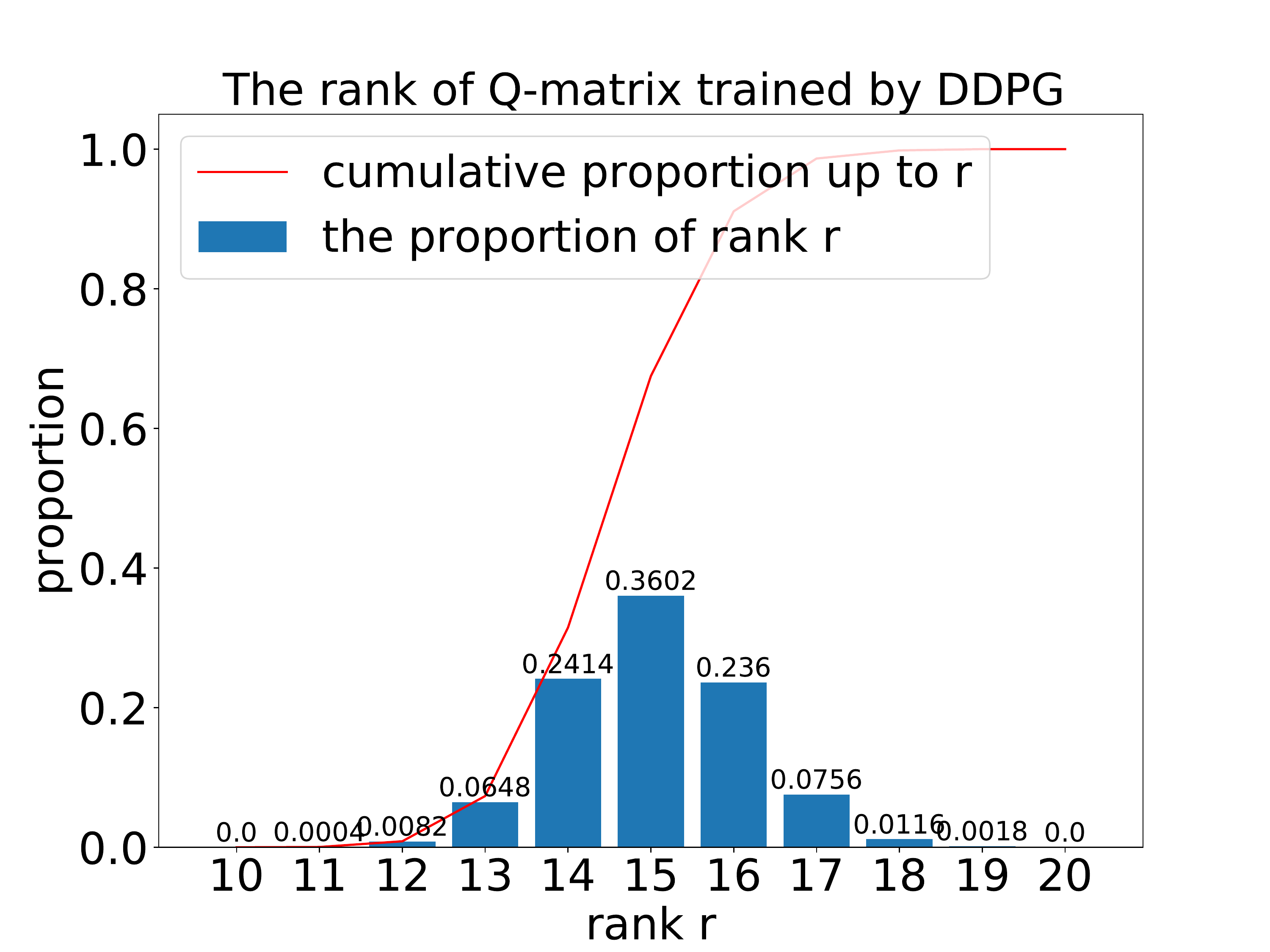}
}
\vspace{-0.3cm}
\caption{\textbf{Distribution of approximate rank of 10k sampled $Q$-matrix} learned by DDPG agent after 1 million training on four MuJoCo environments.
}
\label{figure:rank_distribution}
\end{figure}

To investigate the low-rank structure of $Q$-matrix of DRL agents in continuous control,
we choose three representative DRL agents: DDPG \cite{Lillicrap2015DDPG}, TD3 \cite{Fujimoto2018TD3}, SAC \cite{HaarnojaZAL18SAC}, and we use MuJoCo environments as benchmarks.
We train each algorithm in each environment for 1 million interaction steps;
for each evaluation time point, we record the average return and the approximate rank by calculating the average of 10k sampled $Q$-matrix (i.e., $\{\tilde{Q}_{i}\}_{i=1}^{10000}$ with $|S_{i}|=|A_{i}|=64$ for all $i$, sampled from replay buffer).
In Fig. \ref{figure:low_rank_ph}, we plot the curves of average return and approximate rank of learned $Q$ network (i.e., arank$(Q_{\theta})$.\footnote{For TD3 and SAC which adopt double critics $Q_{\theta_1},Q_{\theta_2}$, we calculate the approximate rank for $Q_{\theta_1}$ since the two are equivalent.}
The results show that 
the rank of sampled $Q$-matrix of all three algorithms decreases during the learning process in all the environments.
This reveals a general but intriguing underlying dynamics of value learning.
Moreover, there shows no clear correlation between the rank and the average return by comparing across different algorithms.
Intuitively, we hypothesize that such phenomena are caused by the integration of function approximation and TD backup,
where the induced generalization and bootstrapping smooth the value estimates.

In Fig. \ref{figure:rank_distribution}, we statistic the distribution of the approximate rank of 10k sampled $Q$-matrix learned by DDPG.
We can find that most $Q$-matrices concentrate at a low rank with the value about 2 to 3 times of the action dimenionality controlled.
This shows that the (near) optimal $Q$-matrix can be low-rank.
In addition, consider the extreme case that $Q$-values are independently influenced by each dimension of action and linear to action, the rank of $Q$-matrix should be the action dimensionality (i.e., $\text{dim}(A)$) since each action's value under some state can be represented by the values of $\text{dim}(A)$ actions.
The results indicate that the (near) optimal $Q$-matrix learned possesses a similarly simple relation between value and action as the extreme case.

\begin{figure*}[t]
\centering
\subfigure{
\includegraphics[width=0.95\textwidth]{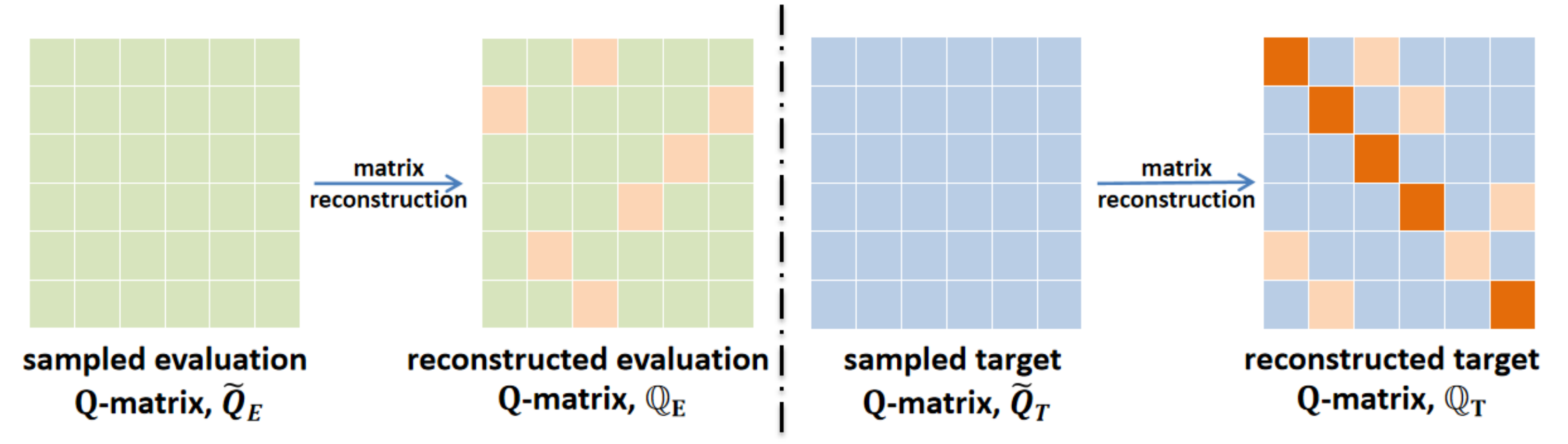}}
\vspace{-0.3cm}
\caption{\textbf{Illustration of two ways of $Q$-matrix reconstruction.}
Consider a batch of transitions $\{(s,a,r,s^{\prime})_i\}$ of size $N$ sampled from replay buffer.
\textit{(Left)}
For the sampled \textit{evaluation $Q$-matrix} (green) $\tilde{Q}_{\text{E}} = \{Q_{\theta}(s,a)\}_{i,j=1,1}^{N,N}$,
the entries (orange) selected (e.g., at random) are erased and then matrix estimation algorithm is performed to obtain the reconstructed $\mathbb{Q}_{\text{E}}$,
which is used to regularize the evaluation network.
\textit{(Right)}
For the sampled \textit{target $Q$-matrix} (blue) $\tilde{Q}_{\text{T}} = \{Q_{\theta}(s^{\prime},a^{\prime})\}_{i,j=1,1}^{N,N}$ where each $a^{\prime} = \pi_{\bar{\phi}}(s^{\prime})$,
the entries are removed and re-estimated in the same way and the diagonal (red) is used as modified target values to approximate.
}
\label{figure:Q_matrix_rec}
\end{figure*}

\subsection{$Q$-Matrix Reconstruction for DRL}
\label{subsection:matrix_rec}

The empirical results above show the dynamics of the approximate rank of $Q$-matrix, i.e., gradually decreasing during the learning process to a low rank.
This reveals the smoothing nature of learning process and the underlying low-rank structure of the desired $Q$ function.
Based on similar empirical results obtained with discrete action space,
Yang et al. \cite{YangZXK20SVRL} propose SVRL algorithm to improve conventional learning process, by conducting matrix reconstruction for target value matrix when performing TD learning of DQN \cite{Mnih2015DQN}.
In the following, we first make a simple extension of the matrix reconstruction for target value matrix to the continuous-action setting.

With discrete (and usually finite) action space,
for each training of value function (recall Eq. \ref{eq:q_td}),
a mini-batch of $N$ transitions $\{(s, a, r, s^{\prime})_{i}\}$ are sampled from the replay buffer.
SVRL \cite{YangZXK20SVRL} uses the $N$ next states and all discrete actions to form the sampled \textit{target $Q$-matrix}.
While it is infeasible to use all actions from a continuous action space,
we make use of the $N$ next actions sampled by target policy $\pi_{\bar{\phi}}$ under corresponding $N$ next states $\{(s^{\prime},a^{\prime})_i\}$ and form an $N \times N$ matrix, denoted by $\tilde{Q}_{\text{T}} = \{Q_{\bar{\theta}}(s^{\prime},a^{\prime})\}_{i,j}$.
As in the left of Fig. \ref{figure:Q_matrix_rec},
by randomly removing a portion of entries in $\tilde{Q}_{\text{T}}$ and then performing Soft-Impute matrix estimation,
we can obtain the reconstructed $Q$-matrix denoted by $\mathbb{Q}_{\text{T}}$,
in which the removed entries are re-estimated by leveraging the underlying rank structure of $Q$-matrix.
Then, we can obtained the modified TD value function learning with target values from $\mathbb{Q}_{\text{T}}$ as shown by the loss function below:
\vspace{-0.1cm}
\begin{equation}
    L_{\text{T}}(\theta)=\mathbb{E}_{s,a,r,s^{\prime} \sim D} \left[Q_{\theta}(s, a) - \mathbb{E}_{a^{\prime} \sim \pi_{\bar{\phi}}(s^{\prime})} \left(r + \gamma \mathbb{Q}_{\text{T}}(s^{\prime},a^{\prime})\right) \right]^{2}.
\label{eq:q_td_rec_target}
\vspace{-0.1cm}
\end{equation}
As proposed in SVRL \cite{YangZXK20SVRL},
the reconstructed value estimates are expected and empirically demonstrated to be better and thus benefit the learning process.
However, we can observe that only the diagonal value estimates in $\mathbb{Q}_{\text{T}}$ is used during updates;
the reconstructed value estimates at randomly removed entries may not be used at all, 
thus being inefficient in leveraging the information of the underlying structure of $Q$-matrix.

To this end, we additionally propose matrix reconstruction on evaluate value matrix for the purpose of making more use of reconstructed value estimates.
As in the right of Fig. \ref{figure:Q_matrix_rec}, for a mini-batch of $N$ transitions, we use the $N$ state-action pairs to form the sampled \textit{evaluation $Q$-matrix}, denoted by $\tilde{Q}_{\text{E}} = \{Q_{\theta}(s,a)\}_{i,j}$. 
Then we can obtain the reconstructed $\mathbb{Q}_{\text{E}}$ and regularize $Q$ network as follows:
\vspace{-0.1cm}
\begin{equation}
    L_{\text{E}}(\theta)=\mathbb{E}_{s,a \sim D} \left[\tilde{Q}_{\text{E}}(s, a) -  \mathbb{Q}_{\text{E}}(s,a) \right]^{2}.
\label{eq:q_td_rec_eval}
\vspace{-0.1cm}
\end{equation}

The above two ways improve value learning process differently.
The former way provides superior target values to approximate by re-estimating values based on rank information 
involved in matrix reconstruction on target $Q$-matrix;
while the latter works more like consistency regularization according to underlying rank structure.

\section{DRL with Uncertainty-aware $Q$-Matrix Reconstruction}
\label{section:ua_lqe}

After the complementary study to prior work on empirical rank structure and reconstructed $Q$-matrix estimates in previous section,
in this section,
we take a further step to reveal the correlation between the approximate rank of $Q$-matrix and value estimation uncertainty.
Based on this discovery,
we propose an uncertainty-aware $Q$-matrix reconstruction method for DRL agents rather than the random removal and reconstruction.

\subsection{Connection between Rank and Uncertainty}
\label{subsection:rank_and_uncer}

\begin{figure}[t]
\centering
\hspace{-0.5cm}
\subfigure[CB@Iter=28]{
\includegraphics[width=0.28\textwidth]{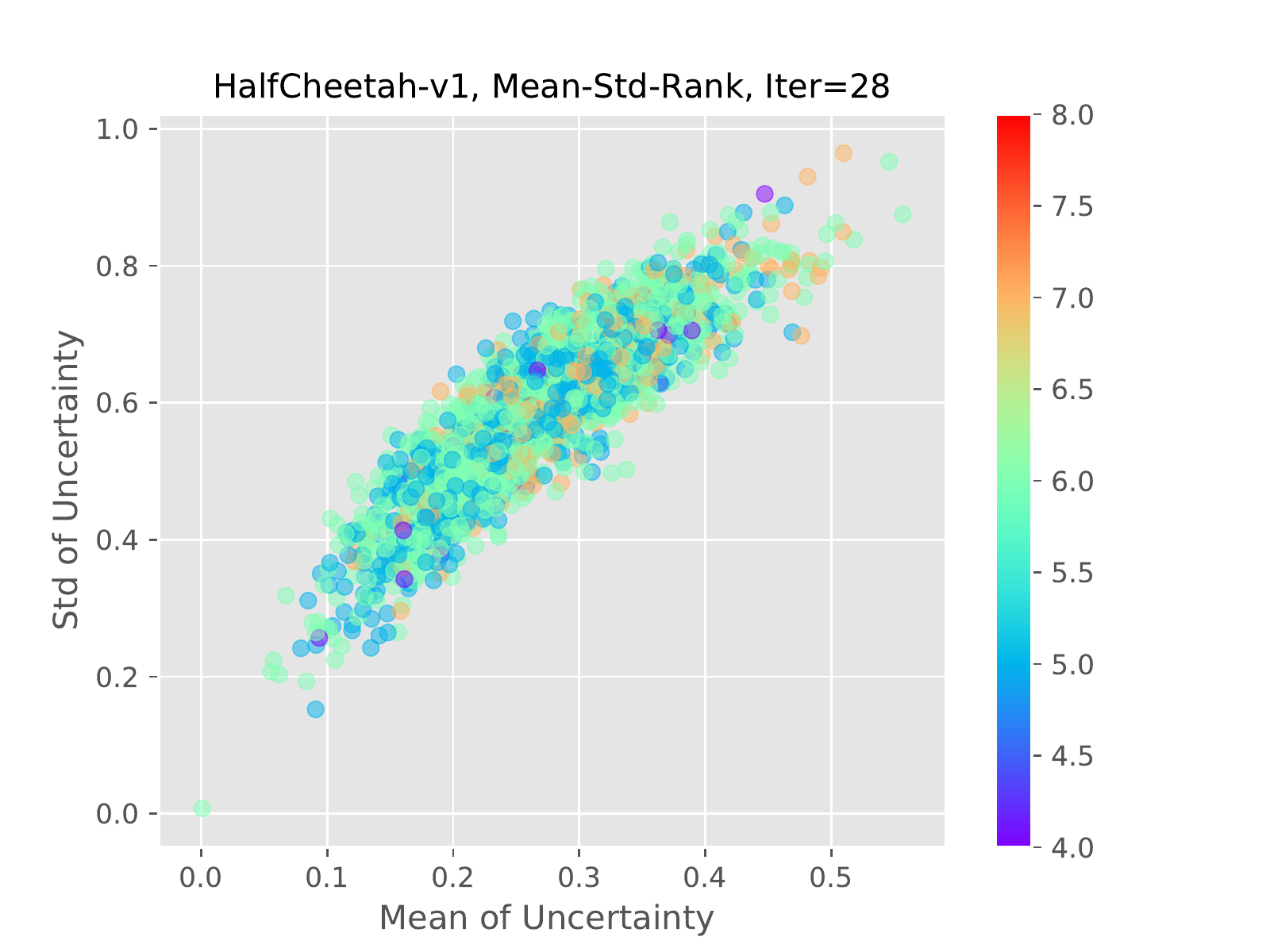}
}
\hspace{-0.85cm}
\subfigure[CB@Iter=38]{
\includegraphics[width=0.28\textwidth]{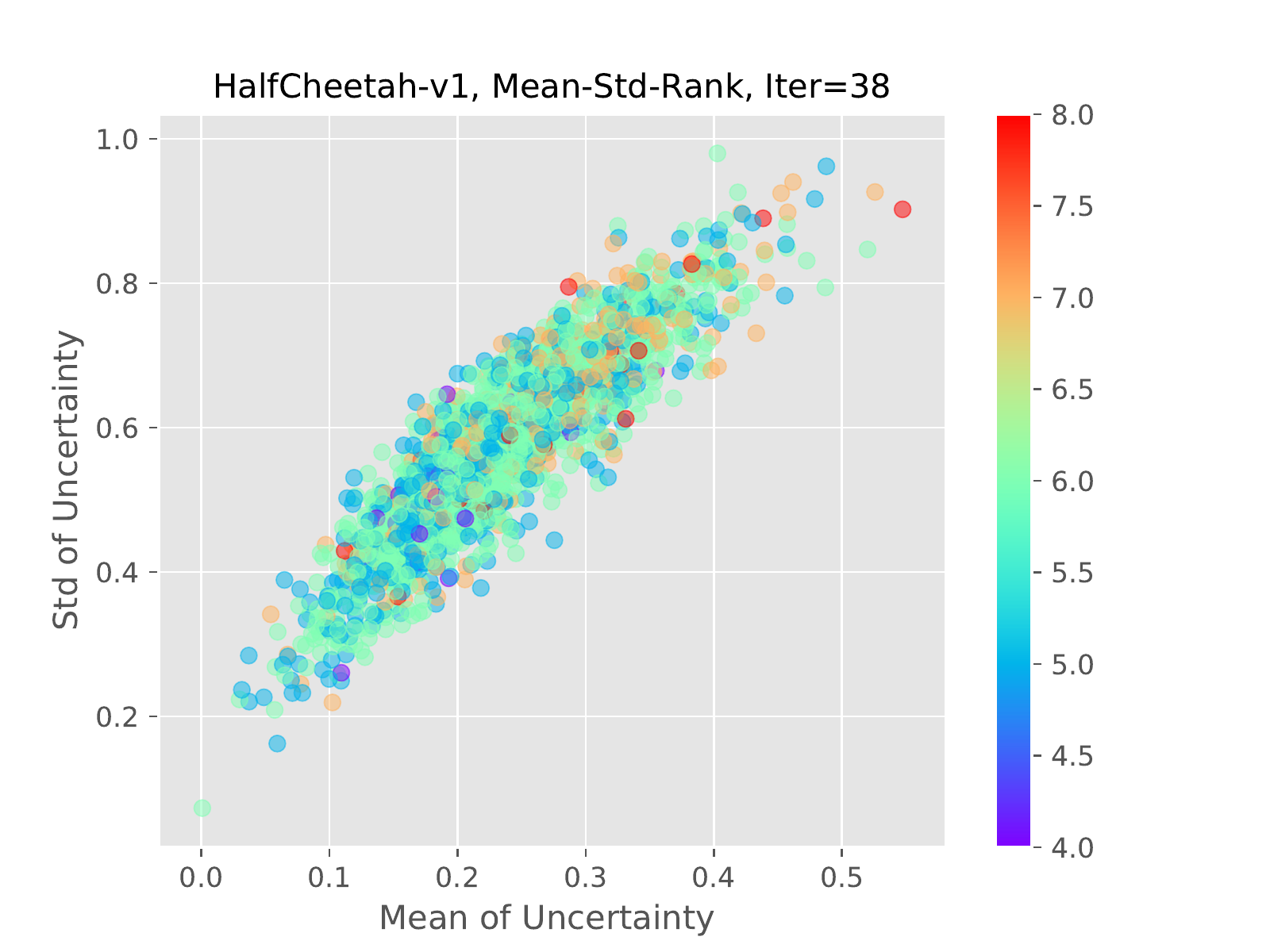}
}
\hspace{-0.85cm}
\subfigure[BB@Iter=28]{
\includegraphics[width=0.28\textwidth]{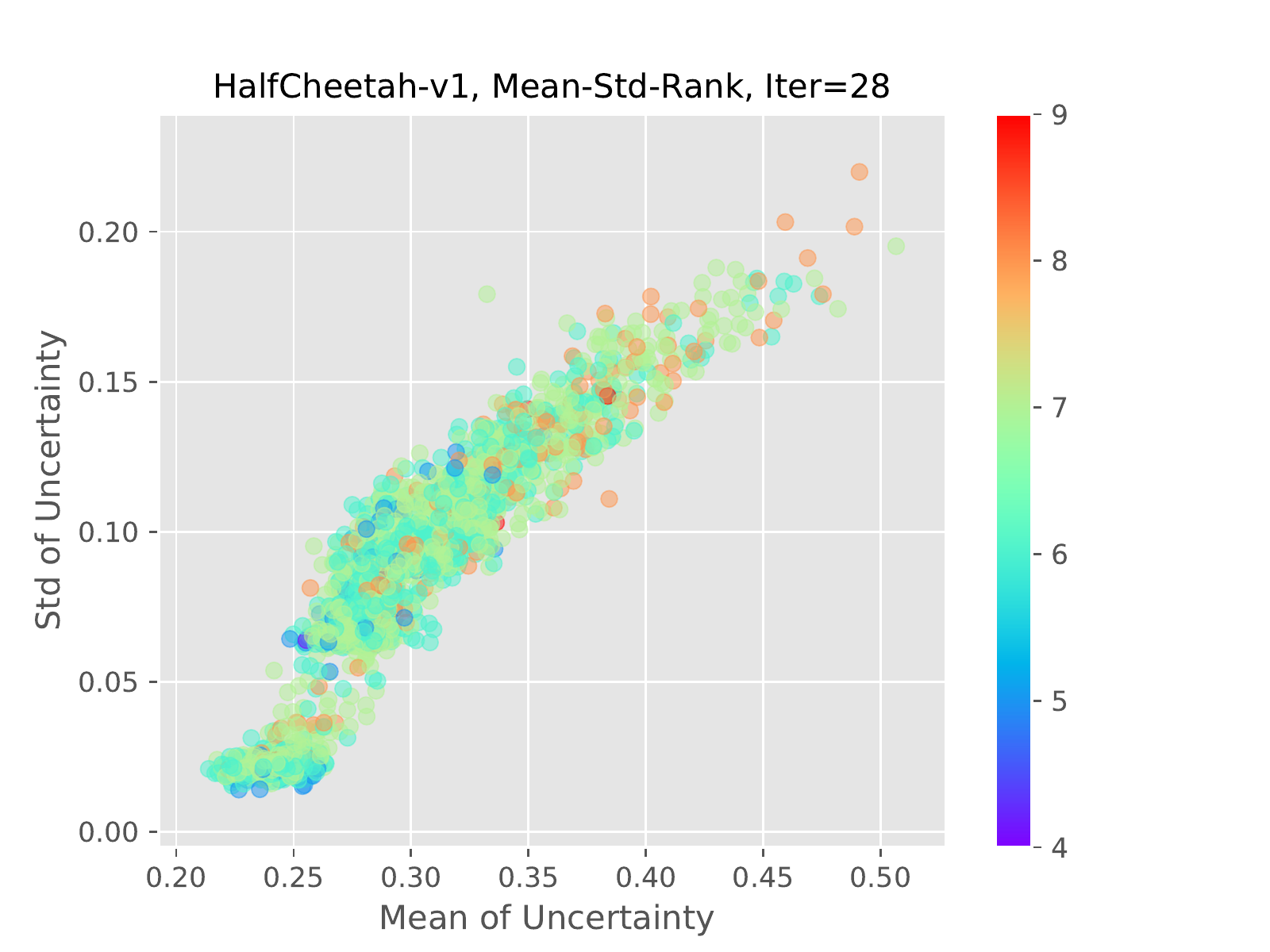}
}
\hspace{-0.85cm}
\subfigure[BB@Iter=38]{
\includegraphics[width=0.28\textwidth]{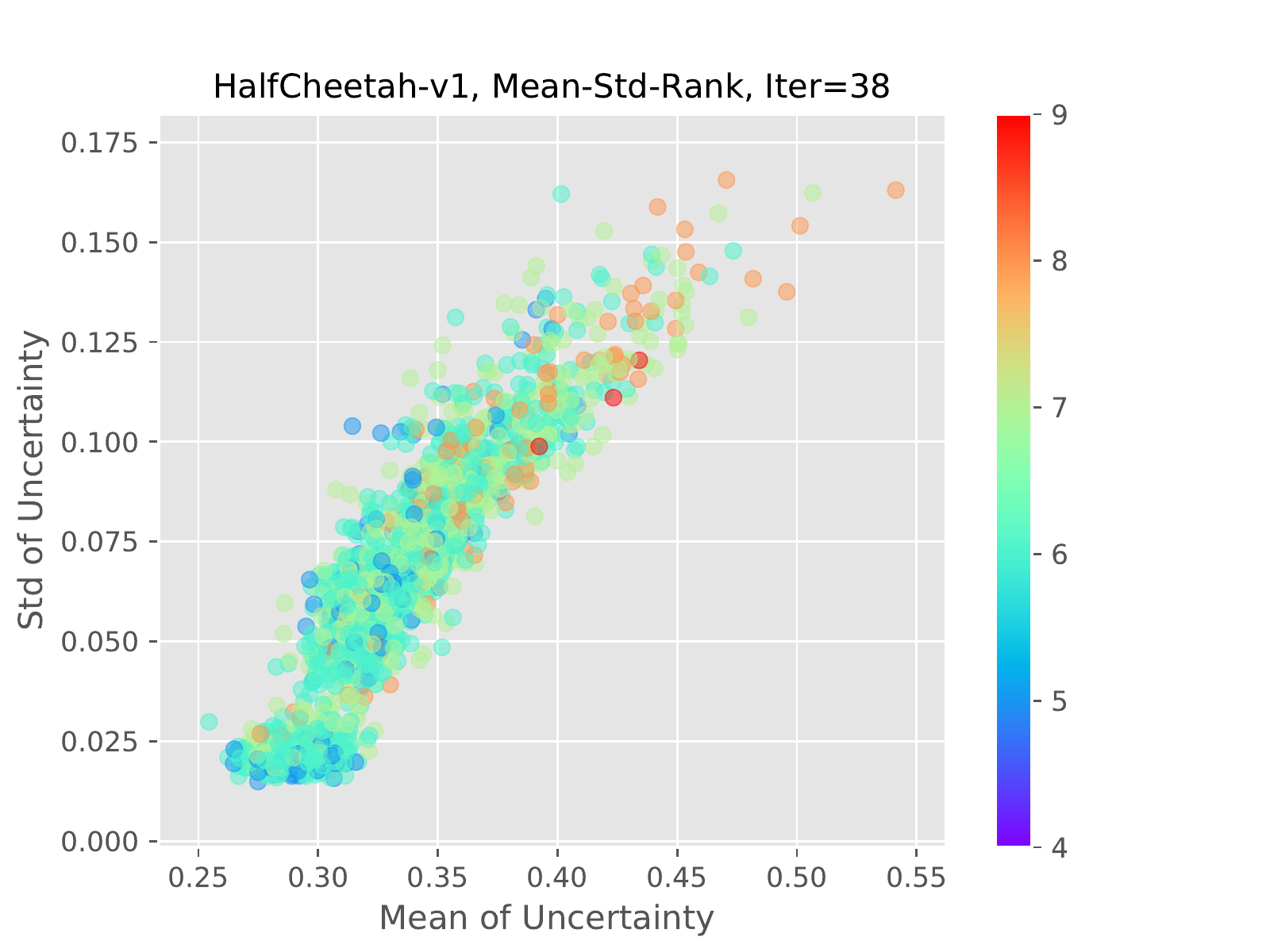}
}
\vspace{-0.3cm}
\caption{\textbf{Illustration of the positive correlation between approximate rank and value estimate uncertainty.}
We use DDPG agent on HalfCheetah for demonstration.
Each point, corresponding to a sampled $Q$-matrix, is colored by its approximate rank with purple and red for the lowest and highest;
and is placed on the 2D plane with the x-axis for the mean of its uncertainty matrix $U(\tilde{Q}_{i})$ and the y-axis for the std.
From \textit{(a)-(d)}, we plot the results in two different iterations for both count-based (CB) and bootstrapped-based (BB) uncertainty quantification methods.
}
\label{figure:rank_uncertainty}
\end{figure}

To investigate the relation between approximate rank of $Q$-matrix and value estimation uncertainty,
we first introduce two techniques for uncertainty quantification:
\textbf{count-based method} \cite{BellemareSOSSM16UnifyCount} and \textbf{bootstrapped-based method} \cite{OsbandBPR16BootstrappedDQN}.

\vspace{-0.3cm}

\subsubsection{Count-based (CB) method}
We denote the sampled times for training of a state-action pair by $N(s,a)$.
For feasibility, original continuous state-action pairs are converted into discrete \textit{Hash} codes \cite{TangHFSCDSTA17countbased} and thus similar pairs are counted together.
We then use the inverse number of sampled times as our first measure of uncertainty \cite{Yang21ExpSurvey}:
$U_{\text{CB}}(s, a) = \frac{1}{N(s,a)}$.

\subsubsection{Bootstrapped-based (BB) method}
Following previous works \cite{OsbandBPR16BootstrappedDQN,CiosekVLH19OAC},
we establish a bootstrapped ensemble of $K$ independent value function estimators $\{Q_{i}\}_{i=1}^{K}$.
For any state-action pair $s,a$,
we obtain the uncertainty $U_{\text{BB}}(s, a)$ by calculating the standard deviation of the value estimates from the ensemble:
$U_{\text{BB}}(s, a) = \sqrt{\frac{1}{N}\sum_{i=1}^N [Q_i (s, a) - \bar{Q} (s,a)]^2}$ where $\bar{Q}$ is the mean of the ensemble.
This is also widely known as \textit{epistemic uncertainty} in the literature of Bayesian exploration.

Next, with the above two uncertainty quantification of value estimates,
we investigate the connection between the approximate rank of sampled $Q$-matrix and value estimate uncertainty.
First, for some time point among the learning process, we sample 100 evaluation $Q$-matrix (similarly as done in Fig. \ref{figure:low_rank_ph} and \ref{figure:rank_distribution}).
For each sampled $Q$-matrix $\tilde{Q}_{i}$,
we obtain its approximate rank;
we also quantify the value estimate uncertainty of each entry in $\tilde{Q}_{i}$, resulting in an uncertainty matrix $U(\tilde{Q}_{i})$, and then calculate the mean and standard deviation (i.e., std for short) of $U(\tilde{Q}_{i})$.
In Fig. \ref{figure:rank_uncertainty},
we draw the scatter plots with the x-axis for the mean of each $U(\tilde{Q}_{i})$ and the y-axis for the std.
Each point in Fig. \ref{figure:rank_uncertainty} denotes a sampled $Q$-matrix and is colored by its approximate rank (from purple to red).
Similar results can also be observed for target $Q$-matrix and for other MuJoCo environments.
Fig. \ref{figure:rank_uncertainty} shows a positive correlation between the approximate rank of sampled $Q$-matrix and value estimate uncertainty.
There also shows a slight trend that this correlation is more obvious as the training marches.

Intuitively, 
when the mean and std are both large, it means that there are many entries with high uncertainty in the $Q$-matrix.
The highly inaccurate value estimates deviate from the underlying structure of true value matrix, thus leading to a high approximate rank.
On the contrary, when the uncertainty is low, in other words the value estimates are relatively certain, the approximate rank is often low since the underlying structure is well approximated.
Now, we reach the important insight that value estimate uncertainty that reflects value approximation error, can be the signifier in finding the target entries which deviate from the desired learning process and prevent the emergence of low-rank structure.
Therefore,
rather than performing at random,
uncertainty-aware removal and reconstruction is aimless and thus can be more efficient in reducing the value approximation error and maintaining the underlying structure of $Q$-matrix, 
finally improving the learning process.

\begin{figure*}[t]
\centering
\includegraphics[width=0.97\linewidth]{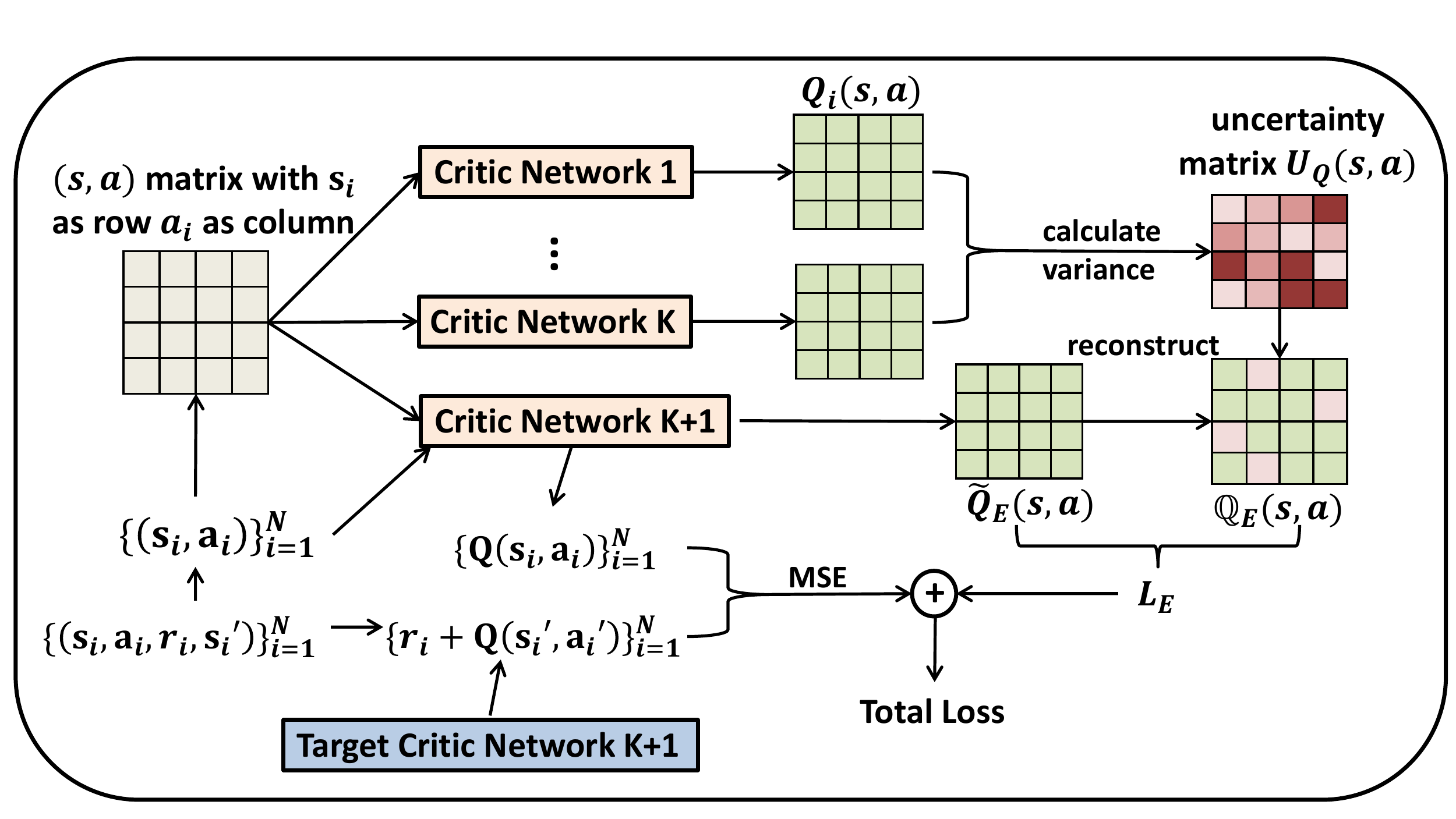}
\vspace{-0.2cm}
\caption{\textbf{Uncertainty-aware $Q$-matrix reconstruction based on} \textbf{bootstrapped-based uncertainty quantification}.
The bootstrapped ensemble of first $K$ critic networks is used to calculate uncertainty, which is used for the selection of re-estimated entries in the $Q$-matrix.
The reconstructed matrix is then used for the learning of the $K+1$ critic network, i.e., the one for primal value estimation and policy learning.
Only matrix reconstruction on sampled evaluation $Q$-matrix $\mathbb{Q}_{E}$ is plot for demonstration.}
\label{figure:bb_ua_matrix_rec}
\end{figure*}

\subsection{Uncertainty-aware $Q$-matrix Reconstruction for DRL}
\label{subsection:algorithm}

Based on the insight revealed in previous paragraph,
in the following we proposed \textbf{U}ncertainty-\textbf{A}ware \textbf{L}ow-rank \textbf{$Q$}-matrix \textbf{E}stimation (\textbf{UA-LQE}) algorithm.

The main idea of UA-LQE is to quantify the uncertainty of $Q$-value estimates during the learning process,
and then perform $Q$-matrix reconstruction as introduced in Sec. \ref{subsection:matrix_rec} with re-estimated entries that are selected according to the quantified uncertainty.
Let us take UA-LQE for \textit{evaluation $Q$-matrix} reconstruction with bootstrapped-based uncertainty quantification for demonstration.
Fig. \ref{figure:bb_ua_matrix_rec} illustrates the process.
We maintain an ensemble of $K$ independent critic networks $\{Q_{i}\}_{i=1}^{K}$ (i.e., $Q$-function approximator) parallel to the major critic network $Q_{K+1}$.
The critic networks in the ensemble are trained in convention according to Eq. \ref{eq:q_td} during the learning process.
For each time of update for the major critic network,
we can obtain the uncertainty matrix of sampled evaluation $Q$-matrix through the ensemble, i.e., $U_{\text{BB}}(\tilde{Q}_{\text{E}})$.
According to $U_{\text{BB}}(\tilde{Q}_{\text{E}})$,
we select the entries of \textbf{top $p$-percent highest uncertainty} for each row (corresponding to each sampled state) of $\tilde{Q}_{\text{E}}$ to remove;
and then we obtain reconstructed evaluation $Q$-matrix $\mathbb{Q}_{\text{E}}$ and conduct the training as Eq. \ref{eq:q_td_rec_eval}.
The processes for target $Q$-matrix reconstruction and for count-based uncertainty quantification are similar which are omitted in Fig. \ref{figure:Q_matrix_rec} for clarity.

Next, we propose a practical implementation of DRL algorithm for continuous control by integrating DDPG as a base algorithm and UA-LQE.
Note that UA-LQE is compatible with most off-the-shelf DRL algorithms.
We use DDPG for an representative implementation and leave the other potential combination for future work.
A detailed pseudo-code of DDPG with UA-LQE, with algorithmic options of sampled $Q$-matrix reconstruction and uncertainty quantification methods,
is provided in Algorithm \ref{alg:E_UALQE}.

\begin{algorithm}[t]
	\small
	Initialize actor $\pi_{\phi}$ and critic networks $Q_{\theta}$ with random parameters $\phi, \theta$,
	and target actor $\pi_{\bar{\phi}}$ and critic networks $Q_{\bar{\theta}}$ with $\bar{\theta} = \theta$, $\bar{\phi} = \phi$
	
	Initialize uncertainty estimator $U_{\omega}(\cdot)$ with parameter $\omega$ (i.e., random Hash table for CB and ensemble of critics for BB) and re-estimation percentage $p$
	
	Initialize replay buffer $\mathcal{D}$, loss weight $\beta$

    \For{Episode e  $\leftarrow$  1  to \text{Max Episode Number}}{
        Obtain the initial state $s_0$
        
    	\For{Timestep t  $\leftarrow$  1  to $ T$}{
    		Select action $a_t \sim \pi_{\phi} (s_t) + \epsilon_{\rm e}$, with exploration noise $\epsilon_{\rm e} \sim \mathcal{N}(0,\sigma)$\\
    		Execute $a_t$ and observe reward $r_t$ and new state $s_{t+1}$\\
    		Store $\{s_t, a_t, r_t, s_{t+1}\}$ in $\mathcal{D}$\\
    		Sample a mini-batch of experience\ $\Omega=\{s_i,a_i,r_i,s_{i+1}\}_{i=1}^{N}$ from $\mathcal{D}$\\
    		\textcolor{blue}{// perform UA-LQE with algorithmic options (refer to Sec. \ref{subsection:matrix_rec})}\\
    		\If(){using Target $Q$-matrix Reconstruction: }{
		        Calculate conventional TD loss $L(\theta)$ as critic loss \Comment{see Eq. \ref{eq:q_td}}\\
		    }
		    \Else(){
		        Establish evaluation $Q$-matrix $\tilde{Q}_{\text{T}}$, remove $p$-percentage entries based on $U_{\omega}(\tilde{Q}_{\text{T}})$ and obtain the reconstructed $Q$-matrix $\mathbb{Q}_{\text{T}}$\\
    		    Calculate $L_{\text{T}}(\theta)$ as critic loss \Comment{see Eq. \ref{eq:q_td_rec_target}}\\
		    }

    		\If(){using Evaluation $Q$-matrix Reconstruction: }{
		        Establish evaluation $Q$-matrix $\tilde{Q}_{\text{E}}$, remove $p$-percentage entries based on $U_{\omega}(\tilde{Q}_{\text{E}})$ and obtain the reconstructed $Q$-matrix $\mathbb{Q}_{\text{E}}$\\
    		    Calculate $L_{\text{E}}(\theta)$, weight it by $\beta$ and add to critic loss \Comment{see Eq. \ref{eq:q_td_rec_eval}}\\
		    }

    		Update $Q_{\theta}$ according to the total critic loss calculated above\\
    		Update $\pi_{\phi}$ with deterministic policy gradient \Comment{see Eq. \ref{eq:DPG}}\\
    		\textcolor{blue}{// update uncertainty estimator (refer to Sec. \ref{subsection:rank_and_uncer})}\\
    		Update $\omega$ accordingly, i.e., update the counts in Hash table for CB and update each critic in the ensemble with $L(\theta_{i})$ for BB\Comment{see Eq. \ref{eq:q_td}}\\
    	}
    }
    %
	\caption{
	DDPG with Uncertainty-aware Low-Rank $Q$-Matrix Estimation \textbf{(DDPG with UA-LQE)}
	}
	\label{alg:E_UALQE}
\end{algorithm}

\section{Experiment}
\label{section:experiment}

In the experiment, we evaluate the efficacy of UA-LQE in several representative MuJoCo continuous control environments.

\subsection{Experiment Setup}
\label{subsection:exp_setup}

We use four MuJoCo environments: \textit{Hopper}, \textit{Walker2d}, \textit{Ant}, \textit{HalfCheetah}, for our experimental evaluation.
We run each algorithm (introduced in the next subsection) for 1 million steps for all four environments,
and evaluate the performance of trained agents every 5000 steps.
For each configuration, we run three independent trials with random environmental seeds and network initialization.

\vspace{-0.5cm}
\subsubsection{Baselines and Variants}
We provide an overview of the properties of the baseline and variants we considered in our experiments in Tab. \ref{table:alg_properties}.
We use DDPG \cite{Lillicrap2015DDPG} as our base algorithm for the implementation of UA-LQE.
We also extend prior algorithm SVRL \cite{YangZXK20SVRL} also based on DDPG for continuous control.
With SVRL, we can evaluate the efficacy of uncertainty-aware $Q$-matrix reconstruction compared with at random.
We also evaluate several UA-LQE variants of different algorithmic options from the perspectives of $Q$-matrix reconstruction losses (i.e., \textbf{E} or \textbf{T}) and uncertainty quantification methods (i.e., \textbf{CB} or \textbf{BB}).

\vspace{-0.5cm}
\subsubsection{Structure and Hyperparameters}
For all algorithms, we use two-layer fully connected networks with 200 nodes and ReLU activation (not including input and output layers) for both actor and critic networks of DDPG base structure.
The learning rate is 0.0001 for actor and 0.001 for critic.
We use a discount factor $\gamma = 0.99$.
The parameters of target networks ($\bar{\theta},\bar{\phi}$) is softly replaced with the ratio 0.001.
After each episode ends, the actor and critic are trained for $T$ times where $T$ here denotes the number of time steps of the episode.
The batch size is set to 64 for both actor and critic mini-batch training and sampled $Q$-matrix (i.e., with the size of 64$\times$64).
The uncertainty-aware re-estimation percentage is set to $p = 20$.
For count-based (CB) method, we use 8-byte Hash codes \cite{TangHFSCDSTA17countbased} for each state-action pair.
For bootstrapped-based (BB) method, we maintain an ensemble of $K=10$ critic networks.
For UA-LQE on evaluation $Q$-matrix,
we weight $L_{\text{E}}(\theta)$ by $\beta=0.1$ when it is added to the total loss.

\begin{table}[t]
	\caption{\textbf{Properties of baselines and variants evaluated in our experiments.} All the algorithms share the DDPG base. N.A. is short for `not applicable'.}
	\centering
	\scalebox{0.9}{
		\begin{tabular}{c|cccc}
			\toprule
			Alg. / Prop. & Reconstruction & Rec. Loss & Selection & Uncertainty ($U_{\omega}(\cdot)$) \\
			\midrule
			\textbf{DDPG} & \XSolidBrush & N.A. & N.A. & N.A. \\
			\midrule
            \textbf{SVRL-E} &	\Checkmark & $L_{\text{E}}(\theta)$ & Random & N.A. \\
            \textbf{UA-LQE-E-CB} &	\Checkmark & $L_{\text{E}}(\theta)$ &	Uncertainty-aware &	Count-based\\
            \textbf{UA-LQE-E-BB} &	\Checkmark & $L_{\text{E}}(\theta)$ &	Uncertainty-aware & Bootstrapped-based\\
            \midrule
            \textbf{SVRL-T} & \Checkmark & $L_{\text{T}}(\theta)$ & Random & N.A. \\
            \textbf{UA-LQE-T-CB} &	\Checkmark & $L_{\text{T}}(\theta)$ & Uncertainty-aware & Count-based\\
            \textbf{UA-LQE-T-BB} & \Checkmark & $L_{\text{T}}(\theta)$ & Uncertainty-aware & Bootstrapped-based\\
			\bottomrule
		\end{tabular}
	}
\vspace{-0.4cm}
\label{table:alg_properties}
\end{table}



\subsection{Results and Analysis}
\label{subsection：results_analysis}

The evaluation results of algorithms are shown in Tab. \ref{table:evaluation_results}.
Complete learning curves are provided in Appendix \ref{appendix:learning_curves}.
We conclude the results from different angles of comparison:
\textbf{1)} Compared with DDPG baseline,
all variants with $Q$-matrix reconstruction (both SVRL and UA-LQE) show an overall better performance (except for some variants in specific environments),
demonstrating the efficacy of $Q$-matrix reconstruction in improving value estimation and thus the entire learning process.
\textbf{2)} UA-LQE variants outperforms SVRL baselines significantly in Walker2d and Ant and show comparable results in HalfCheetah and Hopper.
This shows the effectiveness of uncertainty-aware target entry selection during the process of $Q$-matrix reconstruction.
This reflects the intuition derived from the empirical positive correlation revealed in Sec. \ref{subsection:rank_and_uncer}:
it is more proper to re-estimate the value entries of high uncertainty based on the other ones which are relatively more reliable.
\textbf{3)} The family of evaluation $Q$-matrix reconstruction (\textbf{-E}) performs comparably to the family of target $Q$-matrix reconstruction (\textbf{-T}):
the former performs better in HalfCheetah and Hopper, while the latter is better in Walker2d and Ant.
We hypothesize that this is because two families influence the learning process in different ways (i.e., self-consistency regularization v.s. error reduction in approximation target, as discussed in Sec. \ref{subsection:matrix_rec}).
There remains the possibility to combine of these two orthogonal ways for better results, and we leave the study on such algorithms in the future.
\textbf{4)} For two kinds of uncertainty quantification methods,
bootstrapped-based (\textbf{BB}) variants show superiority over count-based method (\textbf{CB}) in an overall view.
This is consistent to the popularity of bootstrapped-based uncertainty quantification in RL exploration researches \cite{OsbandBPR16BootstrappedDQN,PathakG019ExpDisagree}.
Notably, among all variants, UA-LQE-T-BB achieves significant improvement in Walker2d and Ant, which are of more complex kinematics among the four environments.
This reveals the potential of UA-LQE in improving the performance of DRL agents.

\begin{table}[t]
	\caption{\textbf{Evaluation results of each algorithm (refer to Tab. \ref{table:alg_properties}) in four MuJoCo environments}.
	The results are means and stds of average return after 1 million step over 3 trials.
	Top-1 results in each environment are marked in bold.
	}
	\centering
	\scalebox{1.05}{
		\begin{tabular}{c|cccc}
			\toprule
			Alg. / Env. &	Walker2d & HalfCheetah & Ant & Hopper \\
			\midrule
			\textbf{DDPG} & 1192$\pm$225 & 7967$\pm$439 & 715$\pm$220 & 1348$\pm$339 \\
			\midrule
            \textbf{SVRL-E} &	1458$\pm$238 & 8039$\pm$398 & 443$\pm$221 & 1690$\pm$468\\
            \textbf{UA-LQE-E-CB} &	1835$\pm$348 & 5429$\pm$1020 &	421$\pm$140 &	622$\pm$129\\
            \textbf{UA-LQE-E-BB} &	1716$\pm$503 & 9555$\pm$262 &	707$\pm$203 & \textbf{1751$\pm$480}\\
            \midrule
            \textbf{SVRL-T} & 1863$\pm$296 & \textbf{9819$\pm$392} & 340$\pm$102 & 1721$\pm$422\\
            \textbf{UA-LQE-T-CB} &	2430$\pm$506 & 6764$\pm$346 & 400$\pm$194 & 1029$\pm$202\\
            \textbf{UA-LQE-T-BB} & \textbf{2477$\pm$549} & 8727$\pm$697 & \textbf{1182$\pm$568} & 823$\pm$221\\
			\bottomrule
		\end{tabular}
	}
\vspace{-0.4cm}
\label{table:evaluation_results}
\end{table}

Finally, it remains much space for improving and developing our proposed algorithm UA-LQE.
Our results in MuJoCo is preliminary and obtained without much hyperparameter tuning. A fine-grained hyperparameter tuning is expected to further render the effectiveness of UA-LQE.
Moreover,
it remains space to improve uncertainty quantification with more accurate model and matrix reconstruction with more efficient algorithm.
We leave these aspects for future work.

\vspace{-0.1cm}
\section{Conclusion}
\label{section:conclusion}
In this paper, we demonstrate the existence of low-rank structure of value function of DRL agents in continuous control problem
and introduce two ways of performing $Q$-matrix reconstruction for value function learning with continuous action space.
Moreover, we are the first to empirically reveal a positive correlation between the rank of value matrix and value estimation uncertainty.
Based on this, we further propose a novel Uncertainty-aware Low-rank \textbf{$Q$}-matrix Estimation (UA-LQE) algorithm, whose effectiveness is demonstrated by DDPG-based implementation in several MuJoCo environments.
Our work provides some first-step exploration on leveraging underlying structure in value function to improve DRL,
which can be further studied, e.g., with more advanced and efficient matrix estimation techniques and other re-estimation schemes in the future.

\section*{Acknowledgement}
The work is supported by the National Natural Science Foundation of China (Grant No. 62106172)

\appendix

\section*{Appendix}

\section{More Background on Matrix Estimation}
\label{appendix:bg_matrix_estimation}

Matrix estimation is mainly divided into two types, matrix completion and matrix restoration.
In this paper, we focus on the former. 
We use matrix estimation and matrix completion alternatively. 
Consider a low-rank matrix $M$, in which the values of some entries are missing or unknown.
The goal of matrix completion problem is to recover the matrix by leveraging the low-rank structure.
We use $\Omega$ to denote the set of subscripts of known entries and the use $\hat{M}$ to denote the recovered (also called reconstructed in this paper) matrix.
Formally, the matrix completion problem is formalized as follows:
\begin{equation}
\begin{aligned}
    \min \ \|\hat{M}\|_{\ast}, & \\
    \text{subject to} \ \hat{M}_{i,j}=M_{i,j}, \quad & \forall (i,j) \in \Omega
\end{aligned}
\label{equation:matrix_estimation}
\end{equation}
Here $\| \cdot \|_{\ast}$ is nuclear norm.
The purpose of the above optimization problem is to fill the original matrix as much as possible while keeping the rank of the filled matrix $\hat{M}$ low. 

Some representative methods to solve this optimization problem are Fix Point Continuation (FPC) algorithm \cite{MaGC11FPC}, Singular Value Thresholding (SVT) algorithm \cite{CaiCS10SVT}, OptSpace algorithm \cite{KeshavanOM09OPTspace}, Soft-Impute algorithm \cite{MazumderHT10SoftImpute} and so on. 

\begin{algorithm}[h]
	\small
	Input: Raw matrix $M$, the set of the subscripts of known entries $\Omega$ and the set of the subscripts of unknown (or missing) entries $\bar{\Omega}$\\
	Reset the value of the entries in $\Omega$ of $M$ with $0$ and initialize $M_{old} = M$\\
    Initialize singular filtering divider $\zeta$ and iteration termination threshold $\epsilon$\\
	\For{t  $\leftarrow$  1  to Max Iteration Number}{
	    \textcolor{blue}{// matrix decomposition, singular filtering and matrix reconstruction}\\
	    $U \times S \times V = \text{SVD}(P_\Omega(M_{\text{old}}) + P_{\bar{\Omega}} (M))$\\
	    \textcolor{blue}{// $S$ is the diagonal matrix with singular values $\{s_i\}$ on the diagonal}\\
	    $\lambda = \max_i s_i \ / \ \zeta$\\
		$S_{\lambda} = S - \lambda * E$, where $E$ is the identity matrix\\
		$M_{\text{new}} = U \times S_{\lambda} \times V$\\
		\textcolor{blue}{// check the termination condition}\\
		\If(){$\frac{ {\vert M_{\text{new}} - M_{\text{old}}\vert}_F^2 }{ {\vert M_{\text{old}}\vert}_F^2} > \epsilon$: }{
            Set $M_{\text{old}} = M_{\text{new}}$\\
		}
		\Else(){
		Set Set $\hat{M} = M_{\text{new}}$\\
		break;\\
		}
	}
	Output $\hat{M}$\\
	\caption{Soft-Impute Algorithm
	}
	\label{alg:SF}
\end{algorithm}

In this paper, we use Soft-Impute algorithm for matrix completion, i.e., $Q$-matrix reconstruction. 
A pseudocode for Soft-Impute is shown in Algorithm \ref{alg:SF}.
Define function $P_{\Omega}$ as below:
\begin{equation}
    P_{\Omega}(M) (i, j) = \left \{ \begin{array}{cc}
         &  M_{ij}, \ if\  (i,j) \in \Omega \\
         & 0 , \ if \ (i,j) \not\in \Omega 
    \end{array}
    \right.
\end{equation}
Similar, we define function $P_{\bar{\Omega}}$.
Soft-Impute algorithm completes the missing entries in $\bar{\Omega}$ in an iterative reconstruction fashion.
During each iteration,
the singular matrix $S$ is obtained by SVD decomposition, and then a filtering is performed through subtracting the value $\lambda$ controlled by a filtering divider hyperparameter $\zeta$;
next, the new matrix $M_{\text{new}}$ is reconstructed with the subtracted singular matrix $S_{\lambda}$.
Afterwards, an iteration termination check is performed by comparing whether the difference between the matrices before and after the iteration exceeds the termination threshold $\epsilon$.
When the termination threshold is met or the max iteration number is reached, the algorithm finally outputs the completed matrix $\hat{M}$.

\section{Additional Experimental Details}
\label{appendix:additioanl_experimental_details}

All regular hyperparametyers and implementation details are described in the main body of the paper.

For code-level details, our codes are implemented with Python 3.6.9 and Torch 1.3.1. 
All experiments were run on a single NVIDIA GeForce GTX 1660Ti GPU. 
Our $Q$-matrix reconstruction algorithm is implemented with reference to \texttt{https://github.com/YyzHarry/SV-RL}.

For matrix reconstruction algorithm, i.e., SoftImpute, we set the filtering divider $\zeta=50$, termination threshold $\epsilon= 0.0001$ and Max Iteration Number to be $100$.





\section{Complete Learning Curves of Tab. \ref{table:evaluation_results}}
\label{appendix:learning_curves}

\begin{figure}
\subfigure[Hopper]{
\includegraphics[width=0.475\textwidth]{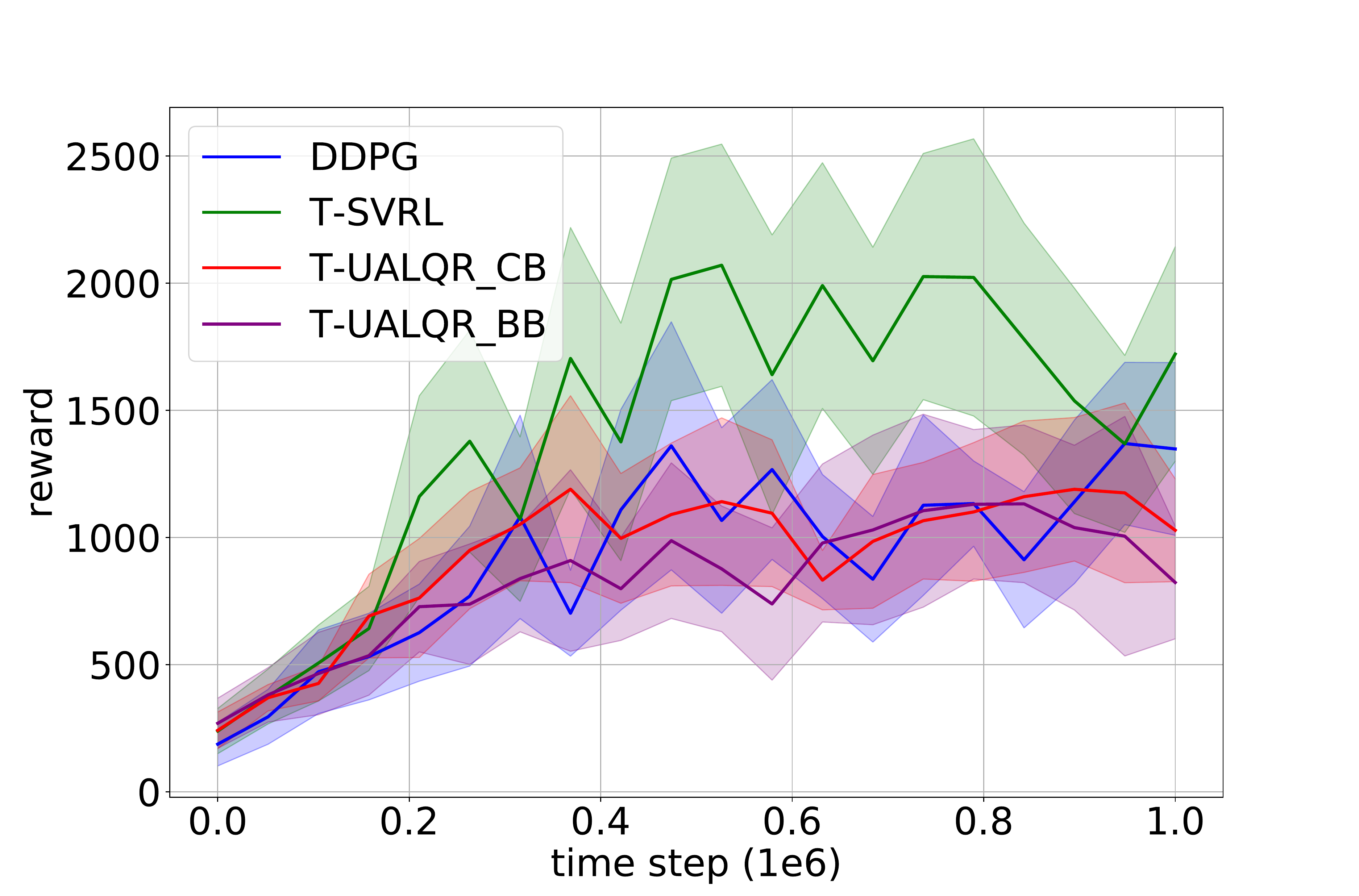}}
\hspace{-0.3cm}
\subfigure[Walker2d]{
\includegraphics[width=0.475\textwidth]{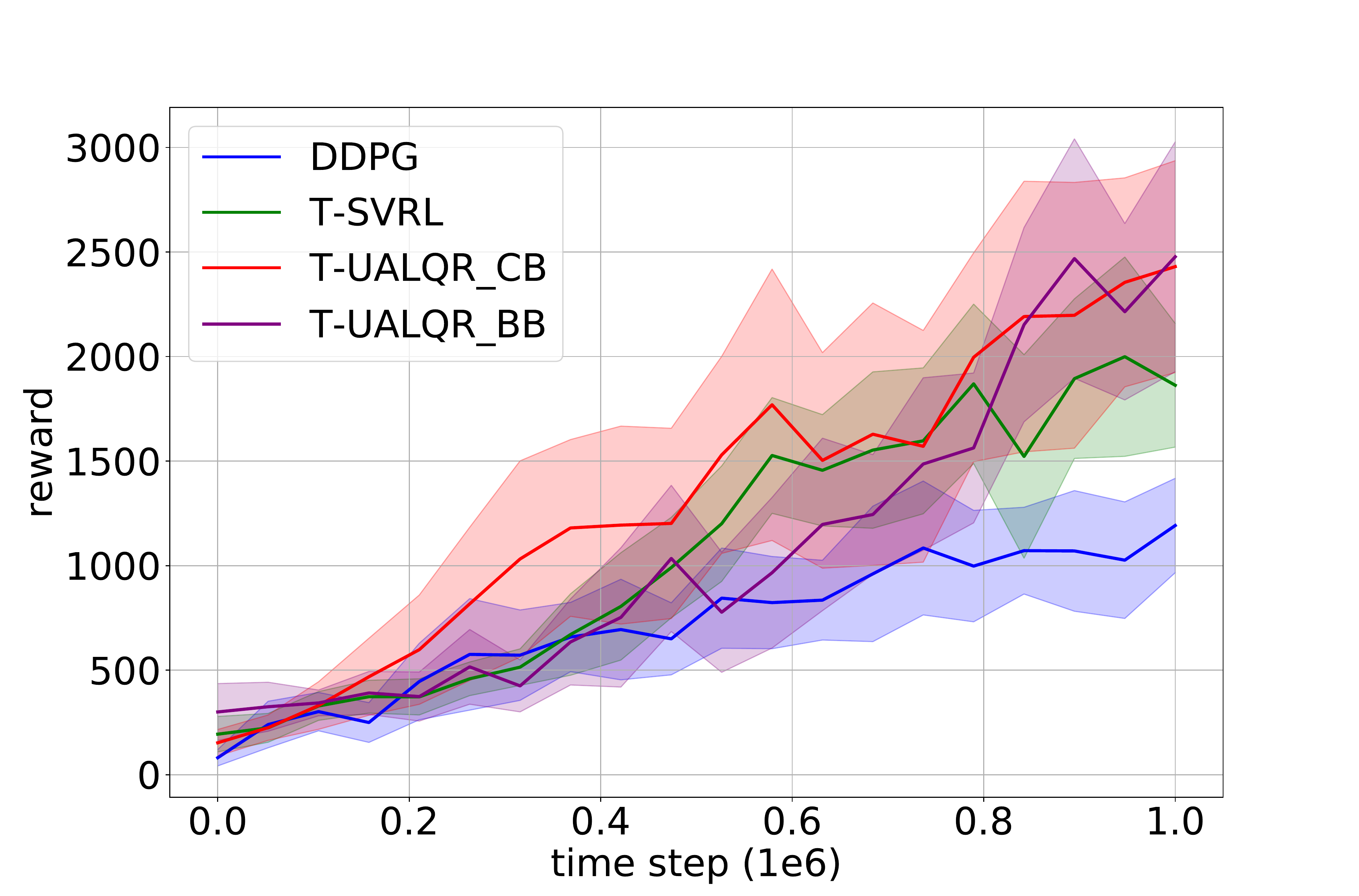}}
\hspace{-0.3cm}
\subfigure[Ant]{
\includegraphics[width=0.475\textwidth]{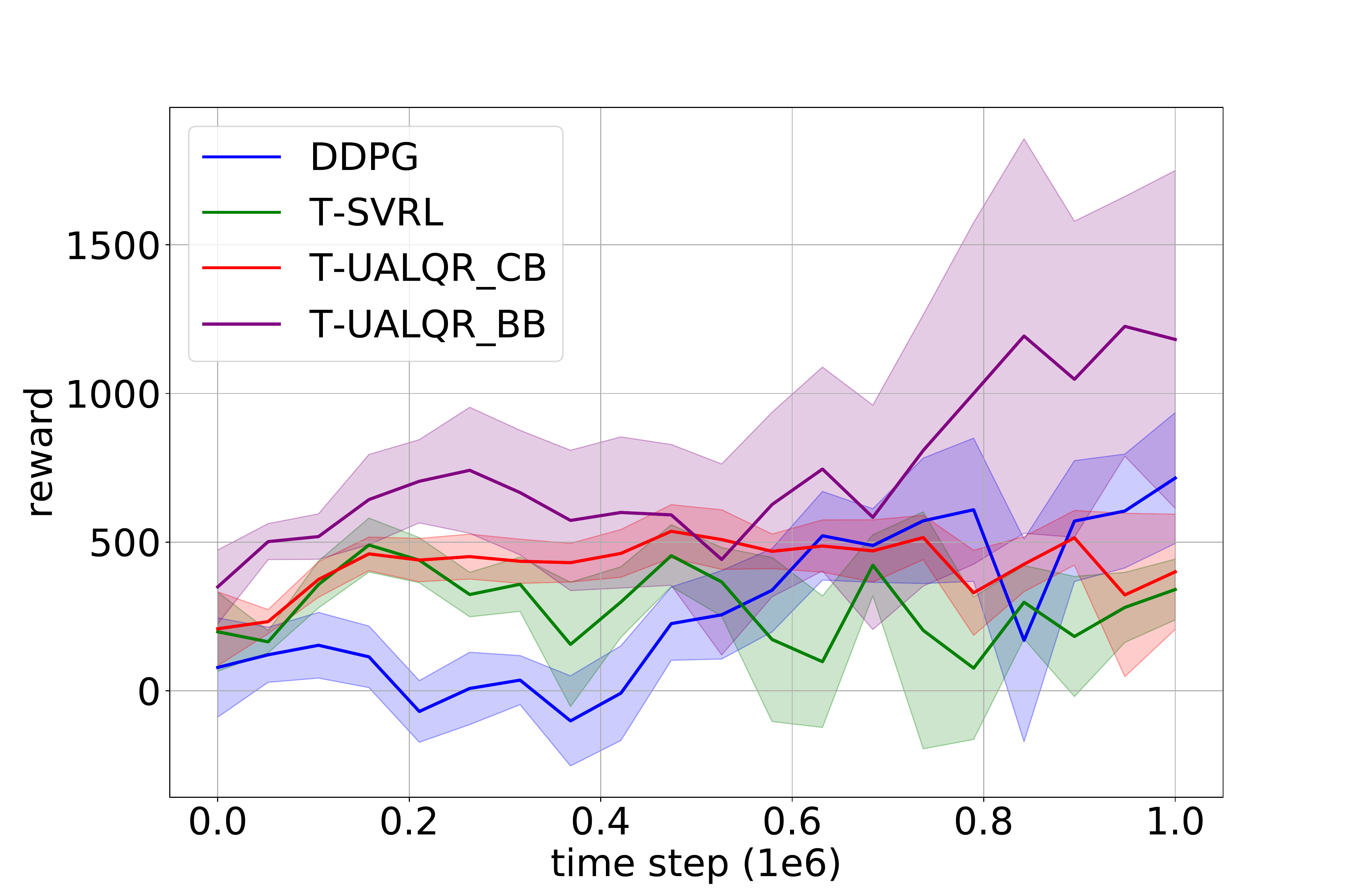}}
\hspace{-0.3cm}
\subfigure[HalfCheetah]{
\includegraphics[width=0.475\textwidth]{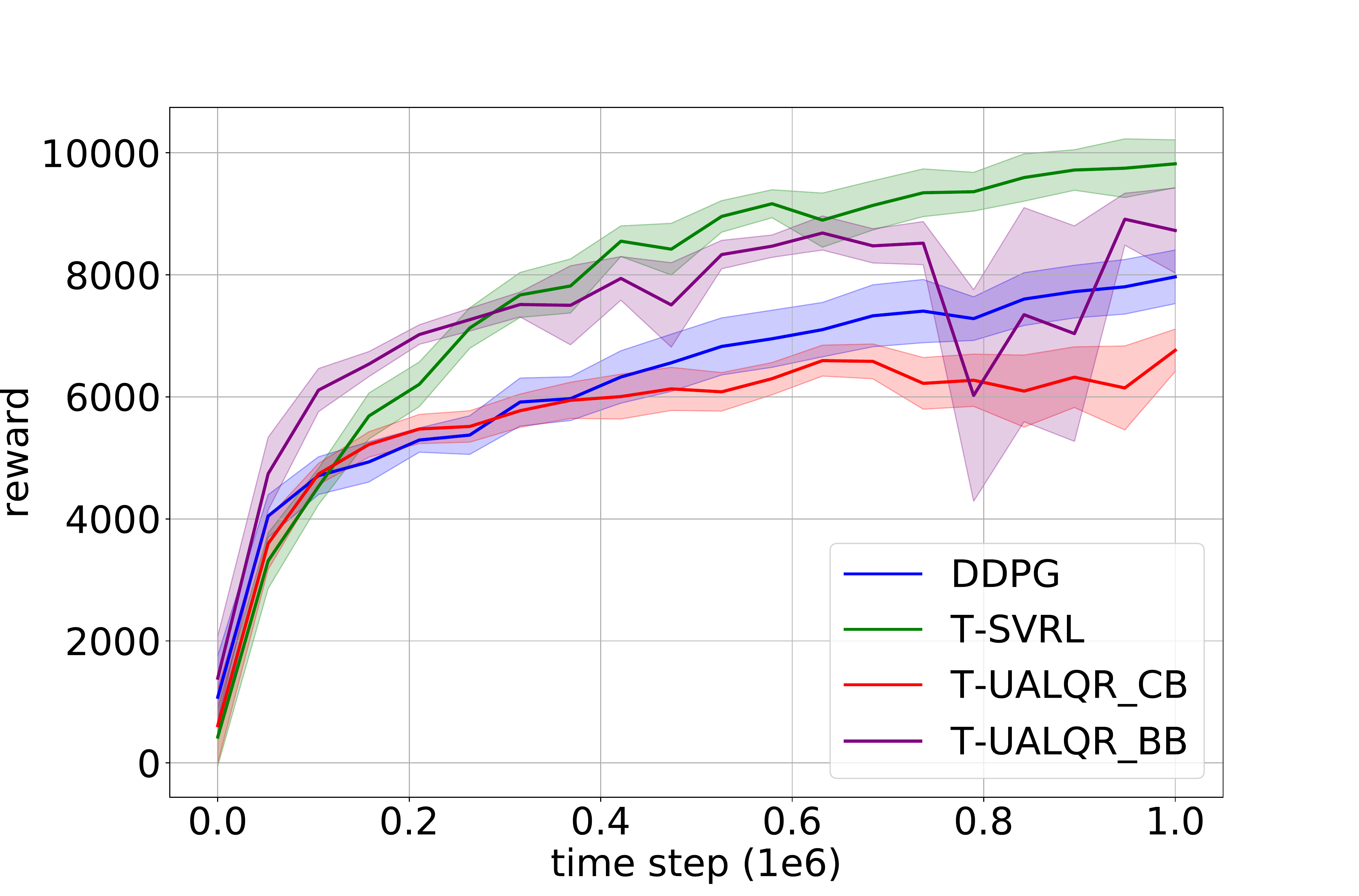}}
\hspace{-0.3cm}
\subfigure[Hopper]{
\includegraphics[width=0.475\textwidth]{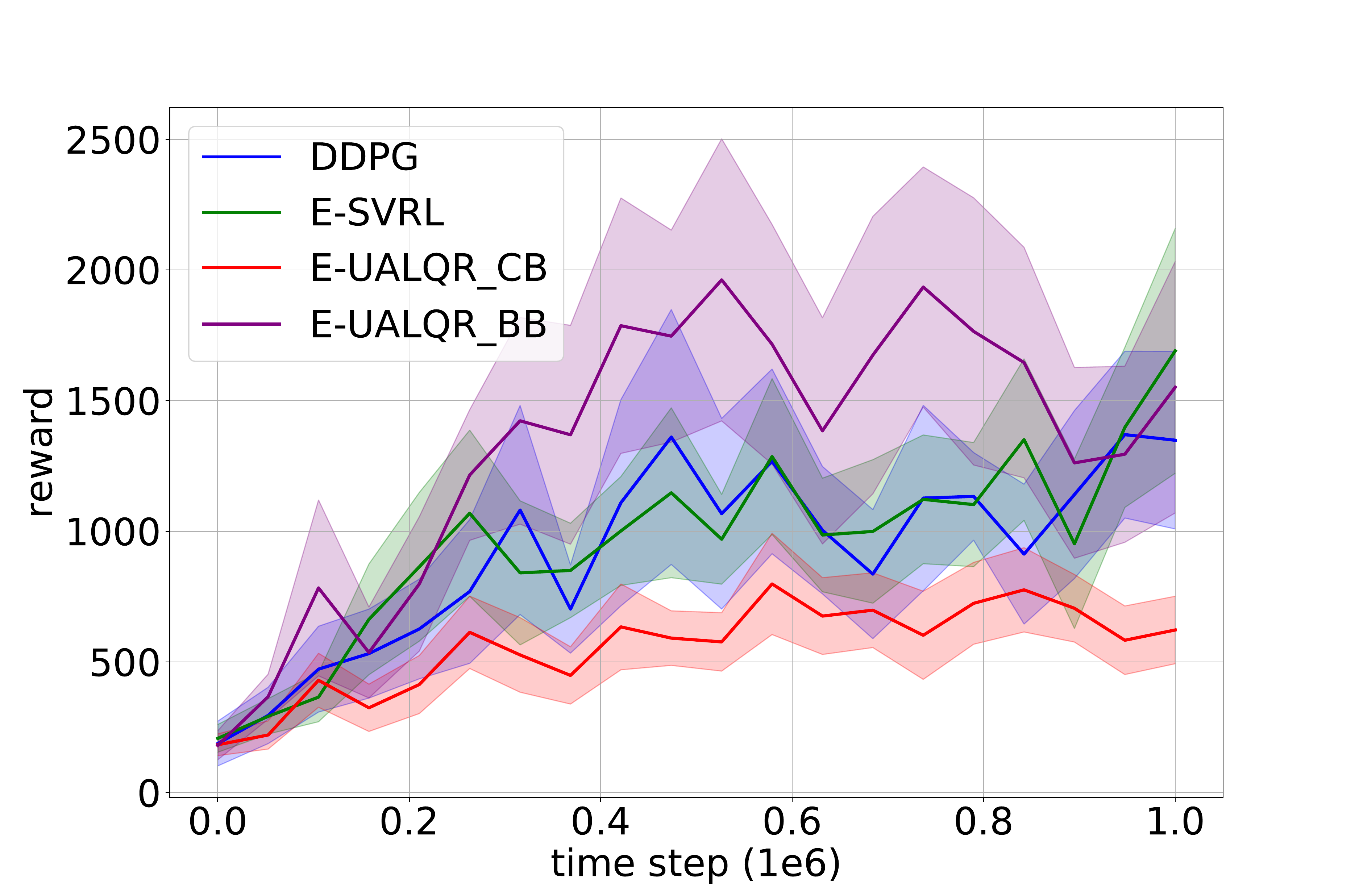}}
\hspace{-0.3cm}
\subfigure[Walker2d]{
\includegraphics[width=0.475\textwidth]{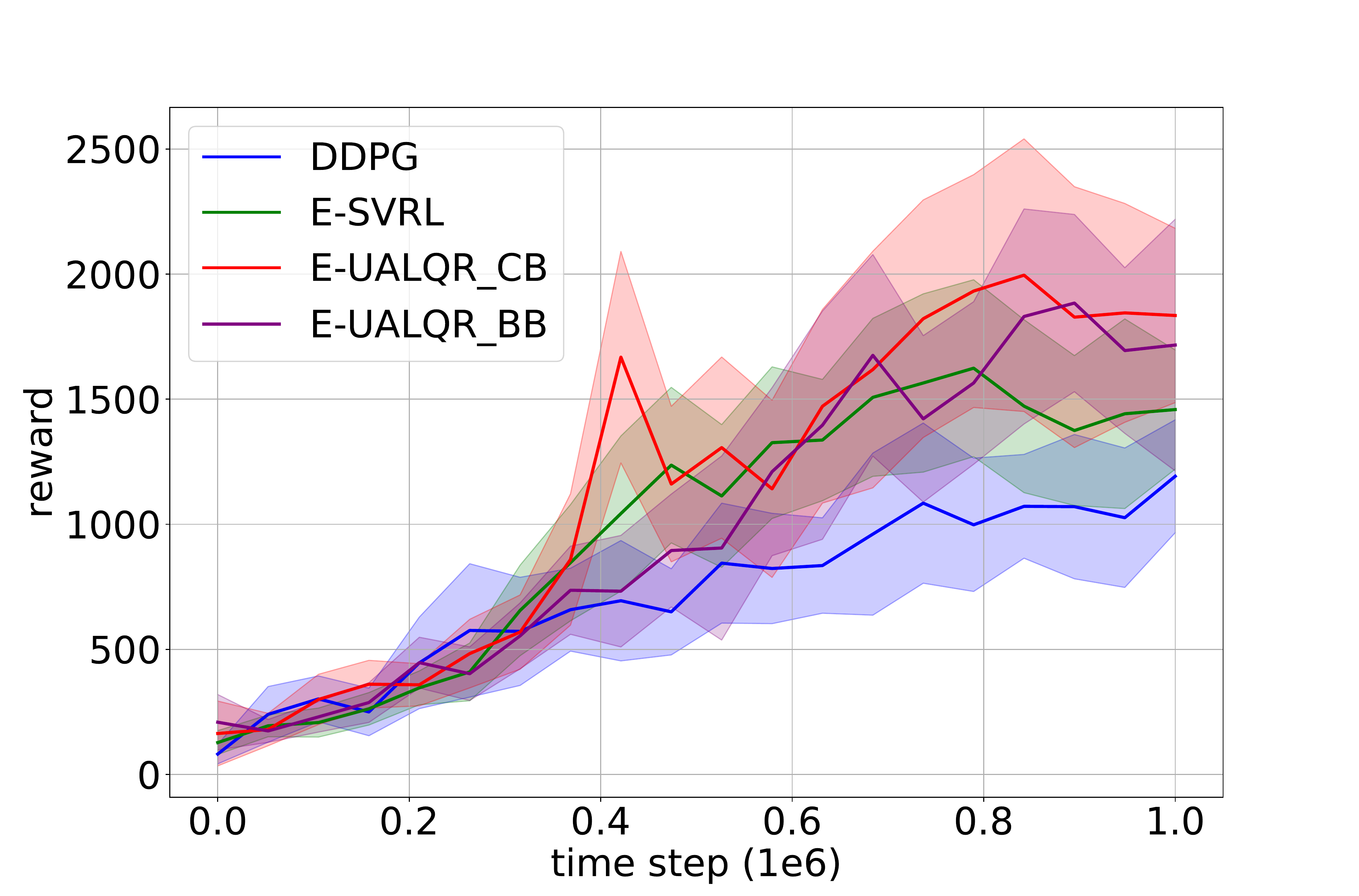}}
\hspace{-0.3cm}
\subfigure[Ant]{
\includegraphics[width=0.475\textwidth]{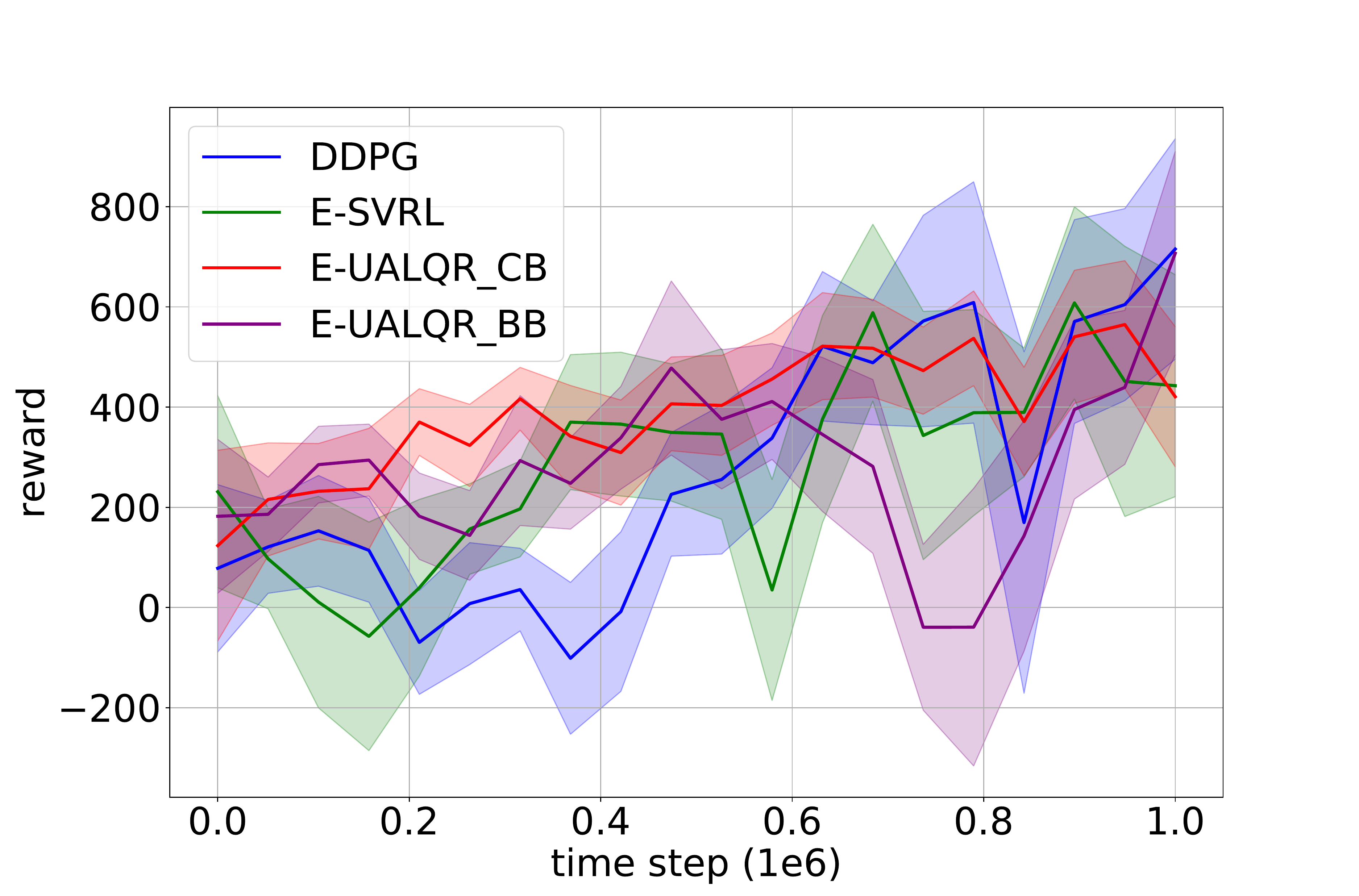}}
\hspace{0.25cm}
\subfigure[HalfCheetah]{
\includegraphics[width=0.475\textwidth]{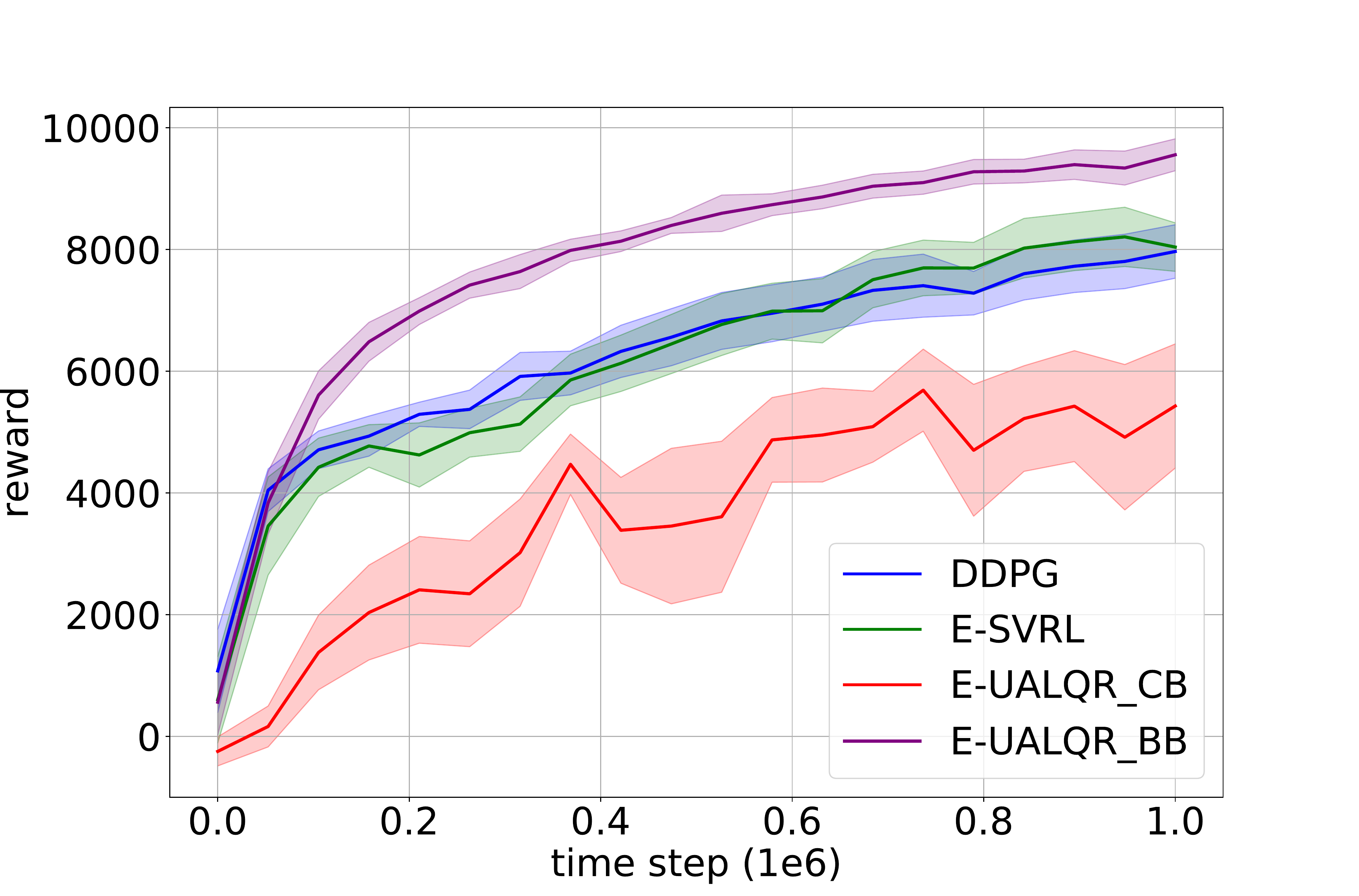}}
\caption{For the reward curve, the first four are the results of reconstruction on the target $Q$-matrix, and the last four are the results of reconstruction on the current $Q$-matrix.
Results are the means and stds over 3 trials.}
\label{figure:evaluation_curves}
\end{figure}

\clearpage
\bibliographystyle{splncs04}
\bibliography{UALQE_ref}

\begin{thebibliography}{10}
\providecommand{\url}[1]{\texttt{#1}}
\providecommand{\urlprefix}{URL }
\providecommand{\doi}[1]{https://doi.org/#1}

\bibitem{BellemareSOSSM16UnifyCount}
Bellemare, M., Srinivasan, S., Ostrovski, G., Schaul, T., Saxton, D., Munos,
  R.: Unifying count-based exploration and intrinsic motivation. In: NeurIPS.
  pp. 1471--1479 (2016)

\bibitem{CaiCS10SVT}
Cai, J., Cand{\`{e}}s, E.J., Shen, Z.: A singular value thresholding algorithm
  for matrix completion. {SIAM} J. Optim.  \textbf{20}(4),  1956--1982 (2010)

\bibitem{CiosekVLH19OAC}
Ciosek, K., Vuong, Q., Loftin, R., Hofmann, K.: Better exploration with
  optimistic actor critic. In: NeurIPS. pp. 1785--1796 (2019)

\bibitem{Fujimoto2018TD3}
Fujimoto, S., v.~Hoof, H., Meger, D.: Addressing function approximation error
  in actor-critic methods. In: ICML (2018)

\bibitem{HaarnojaZAL18SAC}
Haarnoja, T., Zhou, A., Abbeel, P., Levine, S.: Soft actor-critic: Off-policy
  maximum entropy deep reinforcement learning with a stochastic actor. In:
  ICML. pp. 1856--1865 (2018)

\bibitem{HafnerLB020Dream}
Hafner, D., Lillicrap, T.P., Ba, J., Norouzi, M.: Dream to control: Learning
  behaviors by latent imagination. In: ICLR (2020)

\bibitem{Hasselt18DRLDT}
Hasselt, H., Doron, Y., Strub, F., Hessel, M., Sonnerat, N., Modayil, J.: Deep
  reinforcement learning and the deadly triad. CoRR  \textbf{abs/1812.02648}
  (2018)

\bibitem{He21UniformPAC}
He, J., Zhou, D., Gu, Q.: Uniform-pac bounds for reinforcement learning with
  linear function approximation. CoRR  \textbf{abs/2106.11612} (2021)

\bibitem{KeshavanOM09OPTspace}
Keshavan, R.H., Oh, S., Montanari, A.: Matrix completion from a few entries.
  In: {IEEE} International Symposium on Information Theory, {ISIT} 2009, June
  28 - July 3, 2009, Seoul, Korea, Proceedings. pp. 324--328. {IEEE} (2009)

\bibitem{KumarAGL21ImplicitUnder}
Kumar, A., Agarwal, R., Ghosh, D., Levine, S.: Implicit under-parameterization
  inhibits data-efficient deep reinforcement learning. In: ICLR (2021)

\bibitem{Lillicrap2015DDPG}
Lillicrap, T.P., Hunt, J.J., Pritzel, A., Heess, N., Erez, T., Tassa, Y.,
  Silver, D., Wierstra, D.: Continuous control with deep reinforcement
  learning. In: ICLR (2015)

\bibitem{LuoMHCW20I4R}
Luo, X., Meng, Q., He, D., Chen, W., Wang, Y.: {I4R:} promoting deep
  reinforcement learning by the indicator for expressive representations. In:
  IJCAI. pp. 2669--2675. ijcai.org (2020)

\bibitem{LyleROD21OntheE}
Lyle, C., Rowland, M., Ostrovski, G., Dabney, W.: On the effect of auxiliary
  tasks on representation dynamics. In: AISTATS. vol.~130, pp.~1--9 (2021)

\bibitem{MaGC11FPC}
Ma, S., Goldfarb, D., Chen, L.: Fixed point and bregman iterative methods for
  matrix rank minimization. Math. Program.  \textbf{128}(1-2),  321--353 (2011)

\bibitem{MazumderHT10SoftImpute}
Mazumder, R., Hastie, T., Tibshirani, R.: Spectral regularization algorithms
  for learning large incomplete matrices. J. Mach. Learn. Res.  \textbf{11},
  2287--2322 (2010)

\bibitem{Mnih2015DQN}
Mnih, V., Kavukcuoglu, K., Silver, D., Rusu, A.A., Veness, J., Bellemare, M.G.,
  Graves, A., Riedmiller, M.A., Fidjeland, A., Ostrovski, G., Petersen, S.,
  Beattie, C., Sadik, A., Antonoglou, I., King, H., Kumaran, D., Wierstra, D.,
  Legg, S., Hassabis, D.: Human-level control through deep reinforcement
  learning. Nature  \textbf{518}(7540),  529--533 (2015)

\bibitem{Ong15lowrank}
Ong, H.: Value function approximation via low-rank models. CoRR
  \textbf{abs/1509.00061} (2015)

\bibitem{OsbandBPR16BootstrappedDQN}
Osband, I., Blundell, C., Pritzel, A., Roy, B.V.: Deep exploration via
  bootstrapped {DQN}. In: Lee, D.D., Sugiyama, M., von Luxburg, U., Guyon, I.,
  Garnett, R. (eds.) Advances in Neural Information Processing Systems 29:
  Annual Conference on Neural Information Processing Systems 2016, December
  5-10, 2016, Barcelona, Spain. pp. 4026--4034 (2016)

\bibitem{PathakG019ExpDisagree}
Pathak, D., Gandhi, D., Gupta, A.: Self-supervised exploration via
  disagreement. In: ICML. vol.~97, pp. 5062--5071 (2019)

\bibitem{ScherrerGGLG15AMDP}
Scherrer, B., Ghavamzadeh, M., Gabillon, V., Lesner, B., Geist, M.: Approximate
  modified policy iteration and its application to the game of tetris. J. Mach.
  Learn. Res.  \textbf{16},  1629--1676 (2015)

\bibitem{schreck2019retrosyn}
Schreck, J.S., Coley, C.W., Bishop, K.J.: Learning retrosynthetic planning
  through simulated experience. ACS central science  \textbf{5}(6),  970--981
  (2019)

\bibitem{Silver2016Go}
Silver, D., Huang, A., Maddison, C.J., Guez, A., Sifre, L., Driessche, G.,
  Schrittwieser, J., Antonoglou, I., Panneershelvam, V., Lanctot, M., Dieleman,
  S., Grewe, D., Nham, J., Kalchbrenner, N., Sutskever, I., Lillicrap, T.P.,
  Leach, M., Kavukcuoglu, K., Graepel, T., Hassabis, D.: Mastering the game of
  go with deep neural networks and tree search. Nature  \textbf{529}(7587),
  484--489 (2016)

\bibitem{Silver2014DPG}
Silver, D., Lever, G., Heess, N., Degris, T., Wierstra, D., Riedmiller, M.A.:
  Deterministic policy gradient algorithms. In: ICML. pp. 387--395 (2014)

\bibitem{Sutton1988ReinforcementLA}
Sutton, R.S., Barto, A.G.: Reinforcement learning: An introduction. IEEE
  Transactions on Neural Networks  \textbf{16},  285--286 (1988)

\bibitem{TangHFSCDSTA17countbased}
Tang, H., Houthooft, R., Foote, D., Stooke, A., Chen, X., Duan, Y., Schulman,
  J., Turck, F.D., Abbeel, P.: {\#}exploration: {A} study of count-based
  exploration for deep reinforcement learning. In: Guyon, I., von Luxburg, U.,
  Bengio, S., Wallach, H.M., Fergus, R., Vishwanathan, S.V.N., Garnett, R.
  (eds.) Advances in Neural Information Processing Systems 30: Annual
  Conference on Neural Information Processing Systems 2017, 4-9 December 2017,
  Long Beach, CA, {USA}. pp. 2753--2762 (2017)

\bibitem{vinyals2019grandmaster}
Vinyals, O., Babuschkin, I., Czarnecki, W.M., Mathieu, M., Dudzik, A., Chung,
  J., Choi, D.H., Powell, R., Ewalds, T., Georgiev, P., Oh, J., Horgan, D.,
  Kroiss, M., Danihelka, I., Huang, A., Sifre, L., Cai, T., Agapiou, J.P.,
  Jaderberg, M., Vezhnevets, A.S., Leblond, R., Pohlen, T., Dalibard, V.,
  Budden, D., Sulsky, Y., Molloy, J., Paine, T.L., Gulcehre, C., Wang, Z.,
  Pfaff, T., Wu, Y., Ring, R., Yogatama, D., Wünsch, D., McKinney, K., Smith,
  O., Schaul, T., Lillicrap, T., Kavukcuoglu, K., Hassabis, D., Apps, C.,
  Silver, D.: Grandmaster level in starcraft ii using multi-agent reinforcement
  learning. Nature  \textbf{575}(7782),  350--354 (2019)

\bibitem{Yang21ExpSurvey}
Yang, T., Tang, H., Bai, C., Liu, J., Hao, J., Meng, Z., Liu, P.: Exploration
  in deep reinforcement learning: {A} comprehensive survey. CoRR
  \textbf{abs/2109.06668} (2021)

\bibitem{YangZXK20SVRL}
Yang, Y., Zhang, G., Xu, Z., Katabi, D.: Harnessing structures for value-based
  planning and reinforcement learning. In: ICLR (2020)

\end{thebibliography}

\end{document}